\documentclass[lettersize,journal]{IEEEtran}
\usepackage{amsmath,amsfonts}
\usepackage{algorithmic}
\usepackage{algorithm}
\usepackage{array}
\usepackage[caption=false,font=normalsize,labelfont=sf,textfont=sf]{subfig}
\usepackage{textcomp}
\usepackage{stfloats}
\usepackage{url}
\usepackage{verbatim}
\usepackage{graphicx}
\usepackage{cite}
\usepackage{xcolor}
\usepackage{orcidlink}

\begin{document}

\title{Defining `Good': Evaluation Framework for Synthetic Smart Meter Data}

\author{Sheng Chai, Gus Chadney, Charlotte Avery \orcidlink{0000-0001-7812-6795}, Phil Grunewald, Pascal Van Hentenryck, Priya L. Donti
\thanks{This paper was produced by the Centre for Net Zero in collaboration with University of Oxford,  Georgia Institute of Technology, Massachusetts Institute of Technology.}
\thanks{July 16, 2024}
}

\markboth{This is a preprint version and may be submitted to journals for possible publication.}%
{Shell \MakeLowercase{\textit{et al.}}: Definition of Good of Synthetic Smart Meter Data}


\maketitle

\begin{abstract}

Access to granular demand data is essential for the net zero transition; it allows for accurate profiling and active demand management as our reliance on variable renewable generation increases. However, public release of this data is often impossible due to privacy concerns. ‘Good’ quality synthetic data can circumnavigate this issue. Despite significant research on generating synthetic smart meter data, there is still insufficient work on creating a consistent evaluation framework. In this paper, we investigate how common frameworks used by other industries leveraging synthetic data, can be applied to synthetic smart meter data, such as fidelity, utility and privacy. We also recommend specific metrics to ensure that defining aspects of smart meter data are preserved and test the extent to which privacy can be protected using differential privacy. We show that ‘standard’ privacy attack methods like reconstruction or membership inference attacks are inadequate for assessing privacy risks of smart meter datasets. We propose an improved method by injecting training data with implausible outliers, then launching privacy attacks directly on these outliers. The choice of $\epsilon$ (a metric of privacy loss) significantly impacts privacy risk, highlighting the necessity of performing these explicit privacy tests when making trade-offs between fidelity and privacy.

\end{abstract}


\section{Introduction}
\label{sec:intro}
\IEEEPARstart{A}{ccess} to electricity demand time series data is essential to rapid and successful energy transitions \cite{esc_open_sm_data_call}. Researchers, modellers and policymakers need to understand how energy demand profiles are changing, in a system that requires greater real-time optimization of demand and supply on the grid. This data is collected at the household level via smart meters which, in the UK, measure consumption (in kWh) at half-hourly intervals. Yet access to demand data is highly restricted for privacy reasons, which means global energy modelling and policy-making often has to rely on static and highly aggregated data from the past. 

Whilst synthetic smart meter data can help achieve widespread access by allowing us to overcome the privacy barrier, it is critical to ensure that the data produced is indeed privacy-preserving, as well as being `good enough' for its intended purposes. There are numerous investigations into how household characteristics can be inferred from smart meter data in the literature e.g., \cite{beckel2014revealing, radovanovic2022unique}, which suggests smart meter data contains many hidden signatures particular to individual households. This highlights the importance of protecting privacy in synthetic outputs which are trained on real data. To assess whether synthetic data is `good' and `private' enough, an analytical framework with three evaluation concepts is applied: Fidelity, Utility and Privacy. These concepts are commonly applied to evaluate synthetic data in the health sector and other domains with stringent privacy requirements \cite{jordon2022}. The novel contribution of this paper is to translate this framework to the energy sector, specifically with relevance to personal smart meter data. 

We define each of the concepts in this context as follows:

\begin{list}{}
\item{\textbf{Fidelity}: How closely the statistical properties and physical attributes (e.g., the shape, magnitude or frequency of time series features) of the synthetic data match that of real data at the household and aggregated levels.}
\item{\textbf{Utility}: How useful the synthetic data is in addressing particular tasks.}
\item{\textbf{Privacy}: The risk of reconstructing real data or inferring personal data from the synthetic data, and attributing this to individuals.}
\end{list}

Notably, smart meter data is time-series data and is hierarchical in nature. A particular characteristic unique to smart meter data is its ‘spikiness’, which exists in individual as well as aggregated profiles \cite{haben2021review}, e.g., evening peaks in national-level data. Spatio-temporal hierarchies exist in smart meter data when aggregated over time (e.g., daily or weekly) and/or space where, for example, different supply points (e.g., secondary substations) have characteristically different load patterns, owing to weather, network topology and the nature of connected assets. The faithful representation of such features and hierarchies in synthetic data is critical for their fidelity and utility (see e.g., \cite{pang2018hierarchical}).

Machine learning (ML) algorithms used for generating synthetic data, e.g., Variational Autoencoders (VAE)\cite{kingma_vae}, Generative Adversarial Networks (GAN)\cite{goodfellow_generative_2014} and Diffusion-based models\cite{lin_diffusion_2023}, themselves do not guarantee privacy. To mitigate the risk of private data being revealed in synthetic data outputs, many practitioners have resorted to using K-anonymity, or other anonymisation techniques, to protect the privacy of datasets. However, this is not foolproof, as illustrated by examples such as the Netflix prize dataset \cite{netflix-anonymity} and the 1990 US Census Data \cite{sweeney2000simple} cases, where it was possible to reverse-engineer these steps to reveal the full data. 

This paper considers the established concepts of privacy, fidelity and utility (Section \ref{sec:evaluation_frameworks}) and applies them to evaluate the quality of synthetic smart meter data generated using Centre for Net Zero's (CNZ) \footnote{CNZ is an impact-driven research unit, founded by Octopus Energy Group.} Faraday algorithm\cite{faraday_paper}, trained on the publicly available Low Carbon London (LCL) dataset \cite{low_carbon_london} (Section \ref{sec:methodology}). We identify useful and practical means to evaluate whether privacy is preserved in synthetic smart meter outputs by designing privacy-attack experiments in Sections \ref{sec:reconstruction_attack_method} and \ref{sec:MIA_method}. New metrics to evaluate fidelity and utility that account for the unique characteristics of smart meter data are proposed in Sections \ref{sec:fidelity} and \ref{sec:utility}. Given that the disclosure of personal information could occur when the model is subject to overfitting or memorisation of the training data, we evaluate the privacy of generated smart meter data using state-of-the-art techniques introduced in Section \ref{sec:dpsgd} during synthetic data generation. In Section \ref{sec:results} we present the results of our privacy, fidelity and utility experiments.

In Sections \ref{sec:fidelity_privacy_tradeoff} and \ref{sec:utility_privacy_tradeoff}, we further consider the trade-off between fidelity and privacy, as well as utility and privacy, in generating synthetic smart meter data. For this, we apply our recommended fidelity and utility metrics to a synthetic smart meter dataset generated using the Faraday algorithm, whilst varying the level of privacy-preservation in the synthetic data. Finally in Section \ref{sec:summaryofresults}, we use Faraday and the LCL dataset as a case study of how practitioners can apply these learnings in practice within real-world applications.

\section{Background}
\label{sec:background}

Privacy refers to the risk of personal information being revealed. This is especially important for smart meter data which is itself considered sensitive and protected under GDPR. Even a relatively short sequence of half-hourly readings could be sufficient to uniquely identify a given household, and potentially link or infer additional information about it. In this section, we introduce the concept of Differential Privacy (DP), initially proposed by Dwork et al. \cite{dwork2006calibrating, dwork2006our} to quantify privacy risk. DP is considered a state-of-the-art definition of privacy \cite{shahani2023techniques}. For this reason, in Section \ref{sec:methodology} onwards, we consider the impact of implementing DP on privacy attack performance, and its impact on fidelity and utility evaluation, in the context of synthetic smart meter data 
generation. 

\subsection{Differential Privacy}
\label{sec:differential_privacy}
Differential privacy states that the inclusion or exclusion of any single data record should not have disproportionate impact on an algorithm's output. 
More formally, consider two datasets $D_1$ and $D_2$ which differ by only a single record, and let $A$ be a randomized algorithm. DP states that to be considered $(\epsilon, \delta)$-differentially private, for all subsets $S \subseteq \text{Range}(A)$, it must be the case that
%
\begin{equation}
    {\text{Pr}(A(D_1) \in S)} \leq
    e^{\epsilon}{\text{Pr}(A(D_2) \in S)} + \delta.
\end{equation}

Conventionally, DP is used in the setting of querying a database and determining if the result of a query is sufficiently private for it to be released. DP, however, is not without its challenges: 

\subsubsection{DP itself is not an algorithm} It is a mathematical quantification and guarantee of the privacy risk of a dataset, where $(\epsilon, \delta)$ describe how private a dataset is, with smaller $(\epsilon, \delta)$ meaning higher privacy. There is no one defined way of prescribing how DP should be achieved.

\subsubsection{The appropriate values for $(\epsilon, \delta)$ are subjective} Different values are suggested for different use cases \cite{real-world-dp-eps}. There are numerous studies considering `optimal' values for different contexts \cite{dp-choosing-epsilon, how-much-is-enough-dp}.

\subsubsection{There is a trade-off between privacy and fidelity (and by extension, utility)} This makes acceptable values for $(\epsilon, \delta)$ entirely dependent on the use case. For $\delta$, however, common DP practice invokes $\delta \leq 1/N$ \cite{dp-choosing-epsilon} where $N$ is the size of the dataset, as that represents the risk of $\epsilon$ being higher than required being lower than random.


In addition, non-technical stakeholders might not be comfortable with relying on mathematical guarantees alone. As highlighted in the round table by the Financial Conduct Authority on “Exploring Synthetic Data Validation – privacy, utility and fidelity” \cite{synthetic-data-privacy}, participants noted mathematical guarantees such as DP are often insufficient in practice; senior stakeholders would often require explicit checks against privacy risk in addition to mathematical guarantees. To quote a participant in the round table directly: \textit{``[...] customers would feel much safer driving a car that has been thoroughly crash tested over a car that had mathematical ‘guarantees’ of safety built in when it was being developed.''} 

This could be particularly true for smart meter data, where the extent of inferences from the data are not yet well understood. A way of addressing this is to quantify privacy risk through explicit privacy attacks, in addition to relying on differential privacy to quantify the privacy risk of synthetic models and datasets. Jordon et al. \cite{jordon2022} refers to reconstruction attacks, membership inference attack and attribute inference attacks. Therefore, in the following sections (\ref{sec:reconstruction_attack_method}, \ref{sec:MIA_method}), we present explicit privacy evaluation tasks to ensure privacy of synthetic smart meter data in addition to applying differential privacy.

\subsection{Differentially Private Stochastic Gradient Descent}
\label{sec:dpsgd}
Abadi et al. \cite{deep-learning-w-dp} extended DP to the setting of deep learning via the Differentially Private Stochastic Gradient Descent (DP-SGD) algorithm. Deep learning algorithms are generally trained using a procedure called Stochastic Gradient Descent (SGD) \cite{gdt-descent}, which updates model weights in small incremental steps that are based on the gradients of prediction errors on data inputs. The larger the prediction error, the larger the gradient, meaning in the next training step, the model will move further against the direction of the gradient to reduce its prediction error. In practice, SGD utilizes small random batches (mini-batches) of input data due to memory constraints. The intuition behind DP-SGD is that no one single data point should have overly-strong influence over the model weights during gradient descent. To adapt SGD to meet the requirements of DP, \cite{deep-learning-w-dp} proposed the following:
\begin{itemize}
    \item \textbf{Gradient clipping:} For each data point, clip the maximum gradient norm to a `clip value', typically set to 1.0. This limits the amount of influence each data point will have on the weight updates during the training process.
    \item \textbf{Noise Multiplier:} For each batch of data, add just enough noise to mask the largest gradient in the batch (so that by proxy all data will be protected).
    \item \textbf{Privacy-accounting:} During each training step, keep track of the amount of noise added (privacy loss) to approximate $\epsilon$. The privacy accounting operation allows the algorithm to dynamically adjust the amount of noise added (the noise multiplier) to the largest gradient in each batch during gradient descent to achieve the desired $\epsilon$. 
\end{itemize}

An ML algorithm $M$ can be described as $(\epsilon, \delta)$-differentially private if upon the addition or removal of any data point, the probability distribution of the model weights does not change marginally by more than $e^{\epsilon}$ with a probability less than $\delta$. This gives us claims of plausible deniability, in that the weights of the ML model cannot be attributed to any single data point. DP can be implemented in ML models using the DP-SGD algorithm \cite{deep-learning-w-dp} during the training process. DP-SGD however is computationally expensive because it needs to keep track of every single gradient in a batch to identify the training record with the largest gradient and mask it. This expense is lessened with PyTorch's Opacus library \cite{pytorch-opacus} which vectorises this operation.

Throughout this paper, DP-SGD is implemented with PyTorch Opacus \cite{pytorch-opacus} where we set a desired value of $\epsilon$ using the \texttt{target\_epsilon} argument in the \texttt{make\_private\_with\_epsilon} function, which updates the \texttt{noise\_multiplier} on the fly to make sure the observed $\epsilon$ stays within the desired bound.

\section{Evaluation Frameworks}
\label{sec:evaluation_frameworks}

This section describes frameworks for evaluating the fidelity and utility of synthetic smart meter data, as well as two privacy attack frameworks: membership inference and reconstruction attacks. In Section \ref{sec:methodology}, we implement reconstruction and membership inference attacks on a real synthetic smart meter data model as a test for privacy protection in synthetic outputs. Given that smart meter data has unique characteristics which need to be preserved in synthetic outputs, we also outline specific metrics to test for this, and implement these metrics on a real model in Section \ref{sec:methodology}.

\subsection{Reconstruction Attacks}
\label{sec:reconstruction_attack_method}

Reconstruction attacks aim to get the model to output training data that is very similar to the original training data. The success of a reconstruction attack is based on detecting how close synthetic data is to real data. Esteban et al.\cite{esteban2017real} proposed two methods for doing so, using 1) Two-sample Kolmogorov-Smirnov (KS) Test\cite{pratt_kolmogorov-smirnov_1981} and 2) Maximum Mean Discrepancy (MMD) 3-Sample Test\cite{bounliphone_test_2016} to test the null hypothesis of whether the distribution of synthetic data is closer to unseen holdout data than to the seen training data. Failure to reject this null hypothesis would give us the plausible deniability that the model is not regurgitating training data. 
Parikh et al. \cite{parikh2022canary} described Canary Extraction attacks where an attacker is repeatedly querying the model to force it to reveal private data.

Outliers are particularly vulnerable to reconstruction attacks. To investigate the risk of reconstructing outliers, we propose a novel evaluation method called `poisoning attack-based evaluation' inspired by Canary Extraction attacks on large language models. This can be achieved by injecting implausible outliers into a training dataset, and then detecting how many of these implausible outliers were present in the synthetic data output from a generative algorithm. The following steps detail the methodology for the reconstruction attack:

\begin{enumerate}
    \item Construct a sample of artificial outliers, where the number of outliers is small enough to not skew the overall statistics of the training data.
    \item Inject outliers into the training data for the synthetic data model and randomly generate a large sample of synthetic data.
    \item Calculate the pairwise Euclidean distances between the synthetic data points and the artificial outliers. 
    \item Set a `threshold radius' around each outlier based on the risk appetite. This is akin to drawing a circle with a radius equal to the threshold radius around each artificial outlier, and any generated data that falls within this radius renders it `similar enough' to the outlier, such that we consider it to have been successfully reconstructed. To address the problem that acceptable distances are subjective, we develop a novel method for measuring distance which is applicable across different time-series data, described below in Section~\ref{sec:how_close_is_too_close}.
    \item Quantify the \% of injected outliers with at least 1 synthetic data point falling within the threshold radius. If 10 out of the 100 artificial outliers have synthetic data within their radius, we could quantify a privacy leak of about 10\%, i.e. 10\% of outliers were memorised.
\end{enumerate}

\subsubsection{Appropriate distance metric for privacy}\label{sec:how_close_is_too_close}

\begin{figure}
    \centering
    \includegraphics[width=0.6\linewidth]{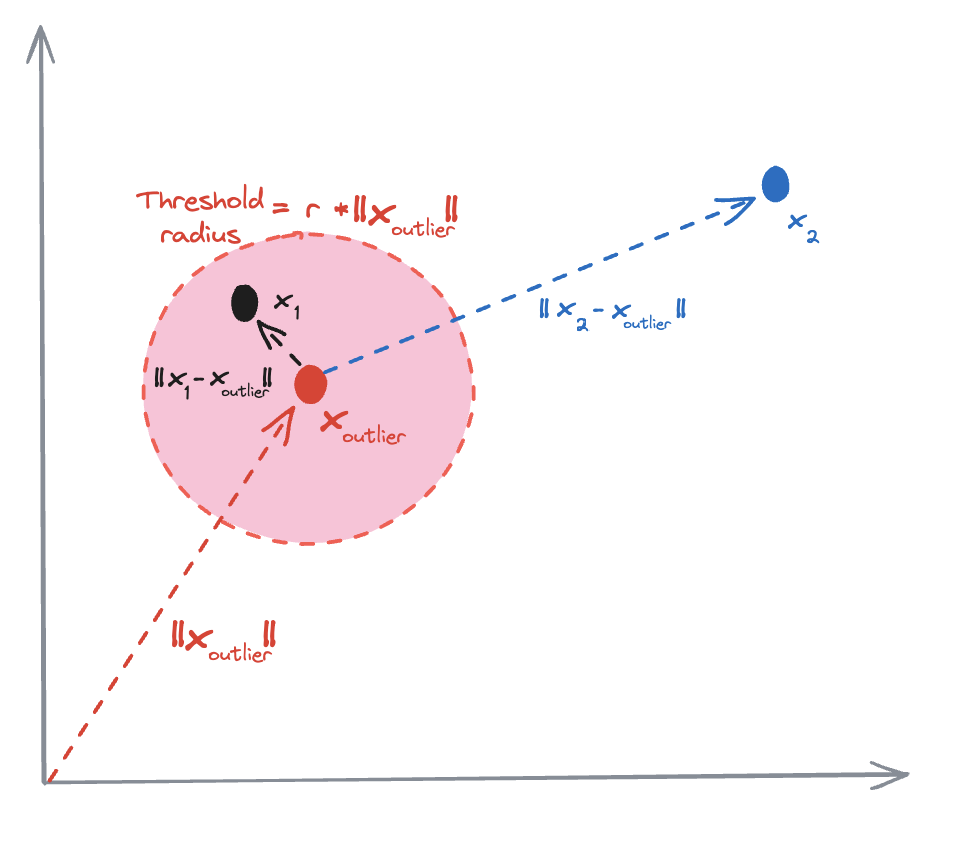}
    \caption{Defining the privacy risk in reconstruction attacks using the ratio to the vector norm. $x_1$, $x_2$ are synthetic data points, $x_{\text{outlier}}$ is an artificial outlier. The threshold radius is the distance boundary within which an outlier is considered to be reconstructed. 
    In this example, $x_1$ falls within the threshold radius; thus the outlier is considered to be reconstructed. Data point $x_2$ does not fall within the threshold radius.}
    \label{fig:radius_vectors}
\end{figure}

Traditionally, in ML, there are two methods of quantifying distances between vectors: 1) Euclidean distance and 2) Distance in unit vectors or cosine similarity. The former measures physically how far two vectors are from each other, whilst the latter measures the angle between the directions in which two vectors are pointing (and thus is magnitude invariant).

For the purpose of reconstruction attacks on time-series data, the threshold radius needs to be somewhat magnitude invariant so that it can be quantified consistently for different lengths of time-series. For example, by virtue of weekly load profiles being longer, Euclidean distances between two weekly load profiles will always be larger than distance between two daily profiles; this makes defining a generic ‘distance threshold’ of how close is ‘close enough’ for reconstruction impossible when using raw Euclidean distance. However, given that the peaks in smart meter load profiles carry information, we cannot completely ignore magnitude and use the angles between unit vectors alone for measuring distance. 

We therefore propose an alternative way of measuring the threshold radius, which is applicable across different time-series data, by taking the radius as a fraction of the norm of the vector describing the outlying data point. Fig \ref{fig:radius_vectors} illustrates how this threshold-distance metric is used for reconstruction. For the reconstruction attack, we take a threshold radius equal to $r \times \|x_{\text{outlier}}\|$, where $\|x_{\text{outlier}}\|$ is the norm of the vector associated with an outlier (in this case, the consumption profile of an injected outlier), and $r$ is the `threshold ratio'. 

To determine the number of compromised outliers, pairwise Euclidean distances between each outlier and each synthetic sample are computed. For each outlier, we obtain the nearest synthetic data point, and if the nearest synthetic data point falls within the specified radius, the outlier is considered to have been reconstructed.

\subsection{Membership Inference Attacks}
\label{sec:MIA_method}
Membership Inference Attacks (MIA) aim to detect whether a given record belongs to the training dataset or not. Successfully predicting whether a record belonged to a training dataset could lead to unintended re-identification and disclosure of personal data. As Hayes et al. \cite{hayes2017logan} illustrated, given an algorithm that recognises criminal faces and the ability to detect if a person belonged to the training data of that algorithm, one could infer that person’s criminal history.

\begin{figure}
    \centering
    \includegraphics[width=\linewidth]{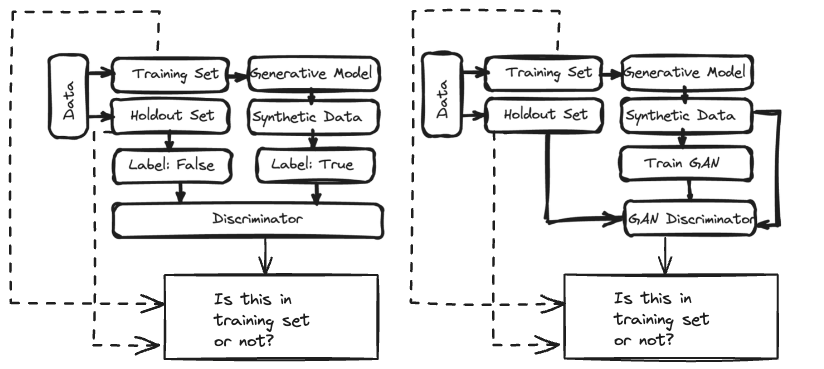}
    \caption{Left: MIA using discriminator with False label for holdout samples and True label for synthetic sample. Right: MIA using a GAN’s discriminator to tell between holdout vs synthetic samples.}
    \label{fig:mia_model}
\end{figure}

MIA in literature has typically been applied to discriminative models (classifiers) \cite{shokri2017membership} by leveraging the fact that ML algorithms tend to output overconfident probabilities on seen samples as opposed to unseen samples. However, a generative algorithm does not output such probabilities that one could leverage to perform MIA. Hayes et al. \cite{hayes2017logan} proposed two methods for performing MIA on generative models which are outlined in Fig. \ref{fig:mia_model}:
\begin{itemize}
    \item Training a discriminator to discriminate between seen and unseen samples.
    \item Training a GAN to copy the generative model to be attacked, and using the discriminator trained in the GAN to discriminate between seen and unseen samples.
\end{itemize}

If the generative model has overfit on (or in the extreme case, is an exact copy of) the training set, we would be able to train a discriminator to differentiate the training and holdout sets. During the attack, we can use the trained discriminator to predict whether a suspected synthetic data point was indeed in the original training set. If the generative model is a 1:1 copy of the training set, and the discriminator is able to tell the difference between synthetic vs holdout data, then for any unseen attack sample, one may be able to successfully establish if this was in the original training data or not.

\subsection{Attribute Inference Attacks}

Attribute Inference Attacks (AIA) aim to infer sensitive information from generated data that the generative model has not seen during training. In the context of smart meter data, sensitive information could refer to attributes like the household income, occupants' gender or religion, or other socio-demographic attributes of households. Such data however is not readily available in the public domain. The Low Carbon London dataset only contains the \texttt{LCLid} as the household identifier with no sensitive information attached. Given the lack of such data to perform AIA on smart meter datasets, we only focus on reconstruction and membership inference attacks for privacy evaluation in this work. We welcome researchers with access to sensitive information alongside smart meter datasets to conduct more research in this area and open-source models that can help quantify the risk of sensitive information being revealed in synthetic smart meter data.

\subsection{Fidelity Metrics}
\label{sec:fidelity}

\begin{figure*}
    \centering
    \includegraphics[width=\textwidth]{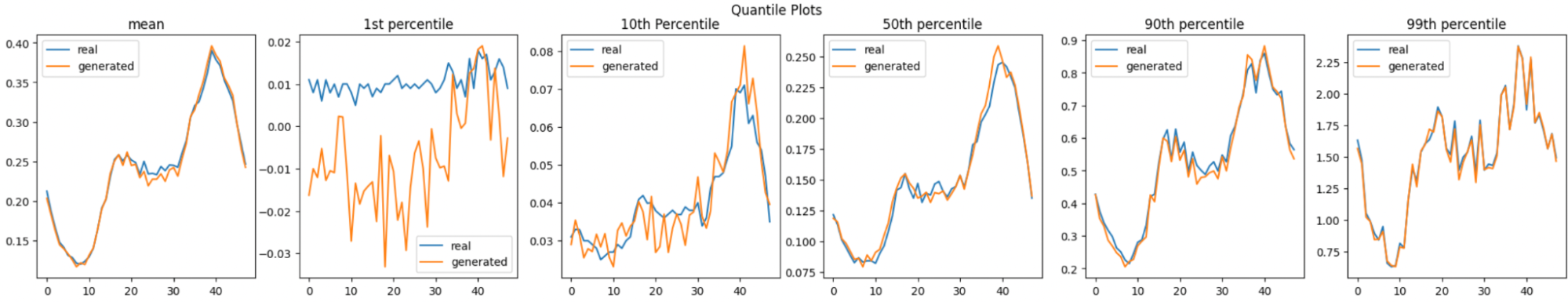}
    \caption{Comparison of consumption quantiles in real smart meter data and synthetic smart meter generated using CNZ's Faraday model \cite{faraday_paper}. Quantile values are calculated at each half-hour window for real and synthetic datasets. Plots show the half-hourly settlement period versus consumption (kWh). 
}
    \label{fig:fidelity_stats}
\end{figure*}

Fidelity refers to how well the statistical properties and physical attributes of synthetic data mirror that of real data \cite{jordon2022}. Tests for fidelity are based on the assumption that if the distribution of synthetic data is a faithful representation of the real data, and if key physical attributes of the real data are preserved in synthetic outputs, then it can be useful for real-life applications. Fidelity metrics which test statistical properties can be categorised into statistical metrics and distance-based metrics. 
Statistical metrics include fundamental properties such as mean, standard deviation or quantiles. More sophisticated approaches include Cramer’s V \cite{tao2021benchmarking}, kurtosis and skewness \cite{diversity-in-synth-ts}. An example comparing the statistical properties between real and synthetic smart meter data generated from Faraday model\cite{faraday_paper} is shown in Fig. \ref{fig:fidelity_stats}.
Distance-based metrics measure the distance of probability distributions between real and synthetic datasets, such as Total Variation Distance (TVD) \cite{tao2021benchmarking}, Maximum Mean Discrepancy (MMD) \cite{mmd-paper}, JS-Divergence \cite{js-divergence}. In \cite{esteban2017real}, Esteban et al. proposed using MMD to assess the fidelity of synthetic data.


\begin{figure}
    \centering
    \includegraphics[width=\linewidth]{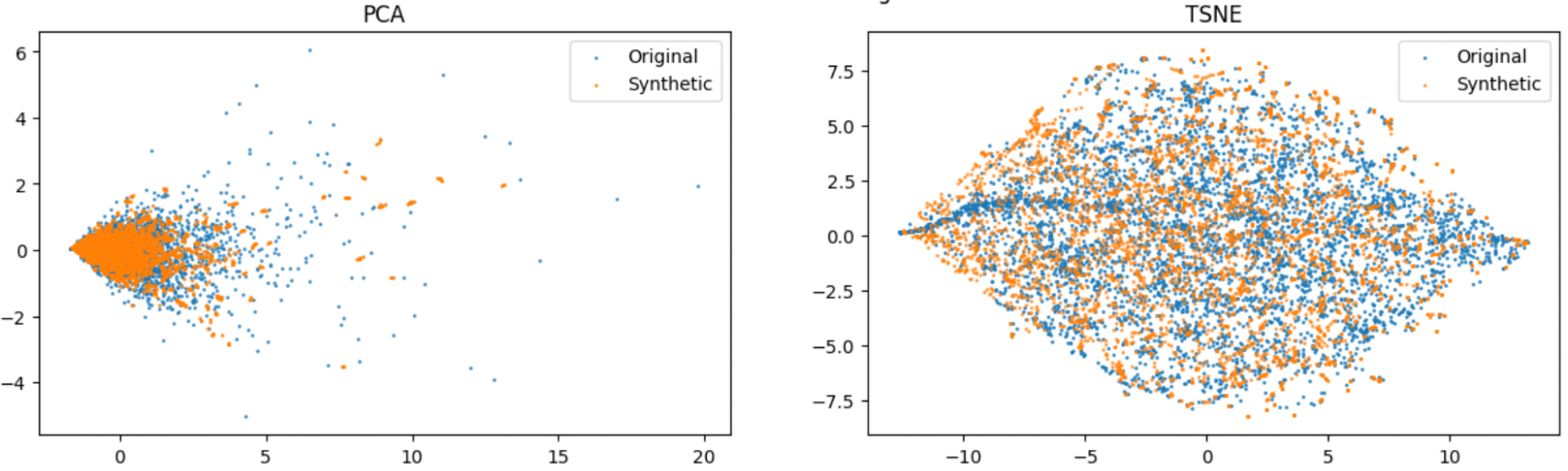}
    \caption{Example of PCA and T-SNE plots of synthetic vs real datasets based on CNZ’s Faraday outputs. 
}
    \label{fig:faraday_pca_tsne}
\end{figure}

Distributions of synthetic smart meter data can also be evaluated qualitatively. Yoon et al. \cite{ts-gan} proposed applying dimensionality reduction techniques e.g., Principal Component Analysis (PCA) and T-distributed Stochastic Neighbour Embedding (T-SNE) to project data onto two-dimensional planes, and plotting the resulting real and synthetic dataset on the same graph to visually inspect the difference in distribution. An example of this applied to Faraday data is in Fig. \ref{fig:faraday_pca_tsne}.

These fidelity metrics, however, do not consider the preservation of the defining characteristics of smart meter data such as temporal-dependence, `spikiness', or spatio-temporal hierarchies. For this reason, we propose the following metrics to better capture characteristics specific to smart meter data:

\subsubsection{Distribution of Autocorrelation Function (ACF) coefficients}
The ACF\cite{hndyman_forecasting} measures how similar a time-series is to a lagged version of itself -- in other words it quantifies the relationship between a given data point (e.g., at 2pm) and preceding data points (e.g., at 1pm). Comparing the ACF coefficients of synthetic versus real data tells us how well the time-series structures are preserved, and helps ensure the preservation of the time-dependence of electricity consumption given that there is an upper limit to consumption. For example, consider a single-occupancy household that may only be able to physically consume 6 kWh per day. If they did all their laundry at 10am (instead of their usual time at 8pm), we would expect a load-shifting behaviour instead of a double spike. 

\subsubsection{Standard statistical and distance-based metrics}
This includes commonly understood statistical metrics like mean, max, quantile, standard deviation etc, and should be calculated across a distribution of profiles at each timestamp (see Fig. \ref{fig:fidelity_stats}). MMD can be used to measure the distances of distributions between real and synthetic data as proposed in \cite{esteban2017real}.

\subsubsection{Distribution of and timing of peaks}
Grid operators are most interested in the timing and magnitude of peaks to identify grid constraints and matching supply with peak demand. It is therefore important for the magnitude and timing of peaks to be preserved in synthetic outputs.

\subsubsection{Distribution of clusters}
Clusters can help us understand how different types of consumers interact with the energy system, and understanding the distribution of clusters can ensure fair representation of `consumer types' in energy system models. Measuring the distribution of clusters helps quantify the joint distribution between all the data points of real and synthetic data. If synthetic data is representative of real data, we would expect cluster distributions of both real and synthetic data to be similar. This can be quantified by first fitting a clustering algorithm on the real dataset. The clustering algorithm is then used to obtain the cluster labels of the real and synthetic dataset, and distributions of the clusters of the two datasets could be compared.

\subsubsection{Similarity of distribution at aggregated levels}
As households are connected to the grid in a hierarchical manner, smart meter data inherently is hierarchical and therefore population dynamics should also be measured. Statistical metrics should be compared not just on a household level, but also on an aggregated level, e.g., across time (e.g., considering total daily consumption), geography, categorical dimensions such as ‘household type’ or sub-population hierarchies e.g., at cluster level. 

This is not an extensive list of valuable fidelity metrics for synthetic smart meter data. Additional metrics may include a measure of demand ramping, i.e., quantifying how rapidly demand rises due to increased loads such as from electric vehicle charging or fleet electric vehicle charging (at the aggregated level). 

\subsection{Utility Metrics}
\label{sec:utility}
Utility refers to how useful synthetic data is when used in real-world applications. Utility is closely related to fidelity, and often fidelity concepts are used to describe the utility of a dataset. This makes sense given that for many applications, synthetic data is only useful if it accurately and fairly represents the real data, as assessed by the fidelity metrics defined in Section \ref{sec:fidelity}. Here, we take the same definitions as \cite{jordon2022} where we consider utility a separate concept that is concerned with the performance of synthetic data when used for a specific task (e.g., forecasting) in place of real data.

In \cite{esteban2017real}, Esteban et al. proposed a framework called `Train on Synthetic, Test on Real' (TSTR) which involves:
\begin{itemize}
    \item `Real' model: Model trained on the real training dataset
    \item `Synthetic model': Model trained on the synthetic training dataset
    \item Both `real' and `synthetic' models are evaluated on an unseen real (test) dataset.
\end{itemize}

The intuition is that for the synthetic data to be useful, the model trained on synthetic data should perform as well as the model trained on the real dataset. The challenge therefore is coming up with relevant example evaluation tasks. These tasks should focus on the qualities and mechanics of smart meter data that we expect the synthetic version to preserve. Given there could be many potential downstream applications of synthetic smart meter data, it is impossible to come up with a set of utility-evaluation tasks that can cover all use cases. That said, we taxonomize downstream uses of smart meter data into two main categories:

\subsubsection{Predictive tasks}
Utility-evaluation tasks should test popular use cases of smart meter data such as classification (e.g., creating `consumer archetypes') or forecasting. A classification task could be based on seen labels during training if the synthetic data is conditioned on these labels. Classes could include Low Carbon Technology (LCT) ownership, or attributes such as household size or property type. Examples of forecasting tasks could include predicting the mean or peak (represented by the 95$^{th}$ percentile) consumption over different time horizons, e.g., intraday (00:00 → 23:00 hrs to predict consumption at 23:30 hrs) and inter-day (e.g. day or week ahead) forecasting. 

\subsubsection{Optimisation/control tasks}
A low-carbon energy system powered by renewables will rely on optimising electricity flows in the grid, especially to balance more variable supply with `spiky' demand.
Synthetic data could potentially play a role in developing optimal power flow models, training reinforcement learning agents for system optimisation, or developing planning models. Utility-evaluation tasks could be set up to test if the outputs of models based on the synthetic data are reasonable.

We note that this is not an extensive list of utility-evaluation tests, and the examples given here involve training ML on the synthetic data. It is important to note that utility-evaluation tasks are not restricted to ML, and what tests should be performed depends on the desired use cases for the dataset. Given the popularity and maturity of prediction models, in the rest of this paper, we focus on testing prediction tasks for utility evaluation only.

\section{Methodology}
\label{sec:methodology}
Here, we outline the methodology used to apply the frameworks introduced in Section \ref{sec:evaluation_frameworks} to a synthetic smart meter data model. We practice a reconstruction attack and MIA for models with and without DP to illustrate the consequences for data privacy in a real-life setting. In the case of fidelity and utility evaluation, we apply metrics 1-5 introduced in Section \ref{sec:fidelity}, and the prediction metrics described in Section \ref{sec:utility}, respectively.

\subsection{Training Dataset}
\label{sec:training_data}
For the purpose of transparency and reproducibility, we use training data from the publicly available Low Carbon London (LCL) dataset \cite{low_carbon_london}, which consists of 30-minute resolution smart metered electricity data for 5,567 households in the greater London area, where data was collected between 2011 and 2014. 

\begin{figure}
    \centering
    \includegraphics[width=\linewidth]{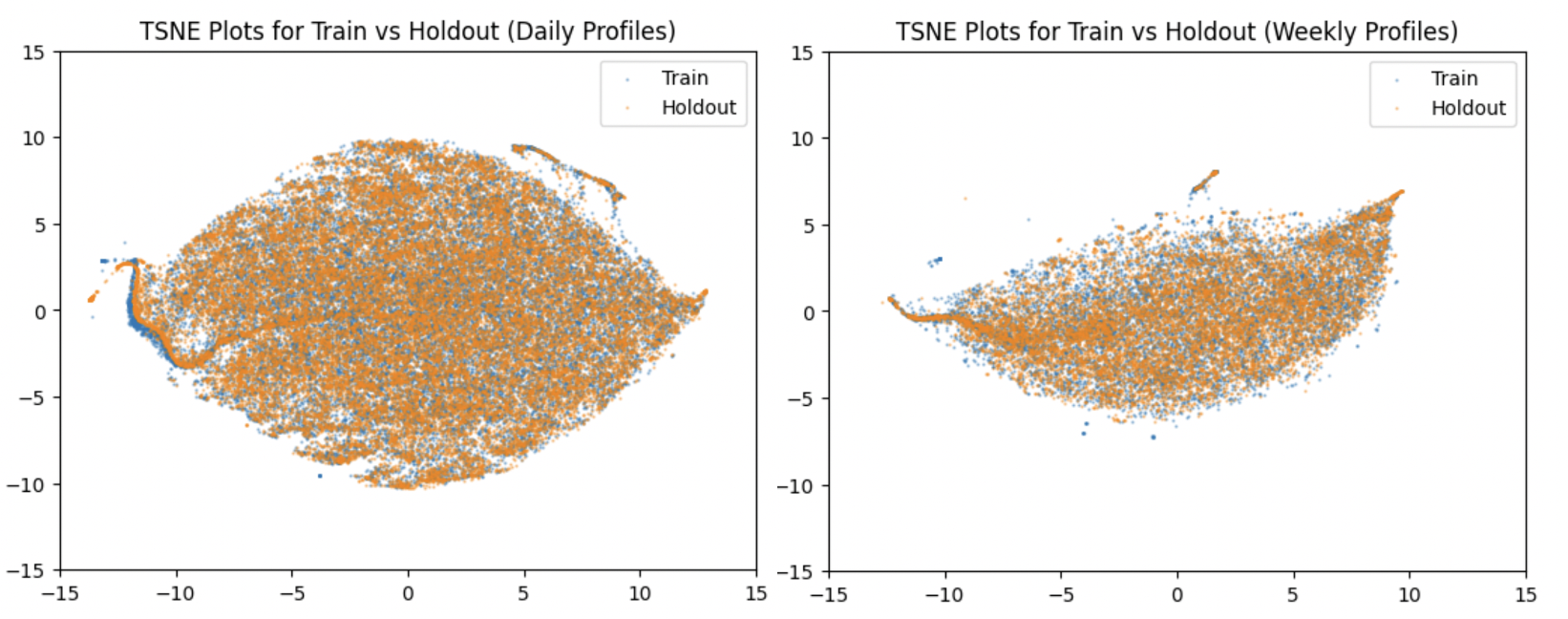}
    \caption{Distribution of Training (Blue) and Holdout (Orange) sets of Daily and Weekly load profiles.}
    \label{fig:low_carbon_ldn_dist}
\end{figure}

The dataset is divided into a `training' (2012 and 2013) and `evaluation' period (2014). Households in the dataset are also randomly split into `training' and `holdout' households. The distribution of the daily and weekly profiles are compared using T-SNE plots in Fig \ref{fig:low_carbon_ldn_dist}. The similarity between the two suggests that the split is sufficiently random.

\subsection{Synthetic Dataset}
\label{sec:synthetic_model}

Synthetic data is generated from the LCL dataset using the CNZ's Faraday model \cite{faraday_paper}. Faraday works by training an VAE to map raw smart meter data into a latent space. A Gaussian Mixture Model (GMM) is then fitted on the distribution of the latent space. During inference, random samples are drawn from the GMM to be decoded by the VAE to retrieve the generated samples. We refer the reader to Section 2 of \cite{faraday_paper} for details on the Faraday model architecture. 

\subsection{Privacy Protection}
Disclosure of personal information could occur when the model is subject to overfitting or memorisation of the training data. We therefore invoke privacy using various methods during synthetic data generation. We use DP-SGD, which acts to prevent overfitting using gradient clipping and noise injection, varying $\texttt{target\_epsilon}$. We also explore whether common regularisation techniques such as L2 penalty \cite{tesauro_scaling_weightdecay} or dropout \cite{srivastava_dropout_2014} could mitigate privacy issues. L2 penalty or weight decay is a common regularisation technique for ML algorithms, where the squared weights are added to the loss term in an effort to penalise overly large weights to limit the influence of any parameter from dominating the model outputs. This is implemented using PyTorch's \texttt{weight\_decay} parameter. Dropout is another common regularisation technique for deep learning architecture to limit overfitting by randomly turning off nodes during training. This has the effect of `thinning’ the neural network to limit the influence of any single node from dominating the model outputs.

All privacy, fidelity and utility frameworks are tested on Faraday outputs, where the generative model is trained with varying levels of differential privacy to evaluate the effectiveness of dataset size, level of differential privacy, and its trade-off on fidelity and utility.

\subsection{Distance-based Reconstruction Attack}
\label{sec:distance_based_reconstruction_attack}
As described by Esteban et al.\cite{esteban2017real}, we consider a reconstruction attack to be unsuccessful if we fail to reject the null hypothesis that the distribution of synthetic data is closer (or as close as) to the unseen holdout data than to the seen data. Using this setup, we calculate pairwise distances of a random sample of synthetic data and their nearest neighbour in the holdout and training set respectively, and compare the distribution of distances using a one-tailed two-sample KS test. 

\subsection{Reconstruction Attack on Outlier Distribution}
\begin{figure}
    \centering
    \includegraphics[width=\linewidth]{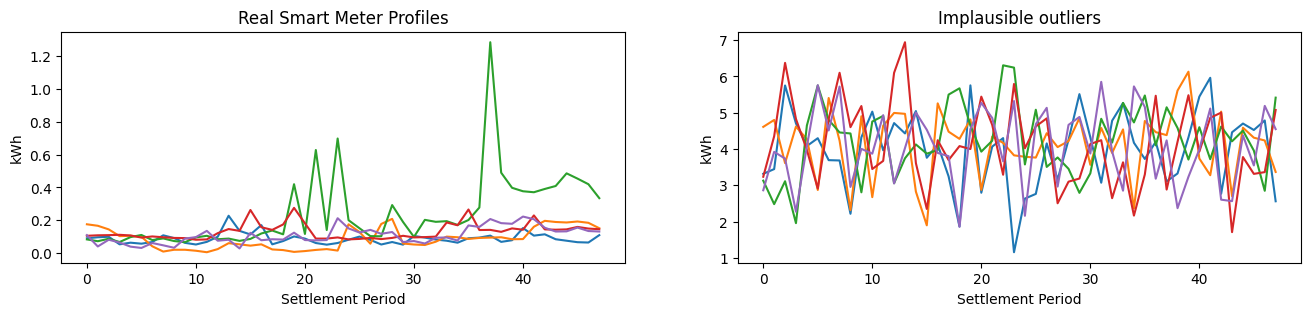}
    \caption{Example of artificial outliers (right) injected into the training data compared to real smart meter profiles (left) from LCL dataset.}
    \label{fig:outlier_profiles}
\end{figure}

The method outlined in Section \ref{sec:reconstruction_attack_method} is implemented by injecting 100 outliers drawn from a normal distribution N($\mu$ = 6kWh, $\sigma$ = 1kWh) into the LCL daily profiles dataset. $\mu$ = 6kWh represents 20-times the population mean of the LCL dataset; when aggregated on a daily basis, this represents a total daily consumption of about 288 kWh, which is implausible for a real household. Examples of these outliers are shown in Fig.~\ref{fig:outlier_profiles}. Note that we choose 100 arbitrarily such that the number of outliers is small enough to not skew the overall statistics of the training data.

We compare the performance of the reconstruction attack when performed on synthetic datasets generated using Faraday model with varying methods of privacy protection: DP ($\epsilon$=1.0, 8.0), L2 penalty and no additional privacy protection.

\subsection{Membership Inference Attack (MIA)}
To test MIA on synthetic smart meter data, synthetic profiles are produced with models that were trained with DP ($\epsilon$=1.0) and without DP, for daily and weekly LCL data respectively. MIA is performed on each dataset using a discriminator model as described in \cite{hayes2017logan} and illustrated in Fig. \ref{fig:mia_model}. This is achieved with the following steps for both the daily and weekly (DP and non-DP) profiles: 
\begin{enumerate}
    \item The holdout dataset is split randomly into two sets: one to train the discriminator model, and one to evaluate during the inference stage. 
    \item Synthetic data is combined with the first holdout set to create a training dataset for the discriminator model. True/False labels were assigned to the holdout set/synthetic data. A discriminator model is then trained to discriminate between the real and synthetic samples.
    \item The ‘attack sample’ is constructed by combining the second holdout set with real LCL training dataset initially used to train Faraday. 
    \item The attack sample is fed into the trained discriminator model to see whether the model could distinguish what samples belonged to the training set. 
    \item Success of the MIA is measured by the precision, where a precision of 1 means the MIA was successful i.e., ‘perfect guess’, and precision equal to 0.5 is equivalent to a random guess on a balanced dataset. 
\end{enumerate}

\subsection{Measuring Fidelity}
\begin{figure}
    \centering
    \includegraphics[width=0.5\linewidth]{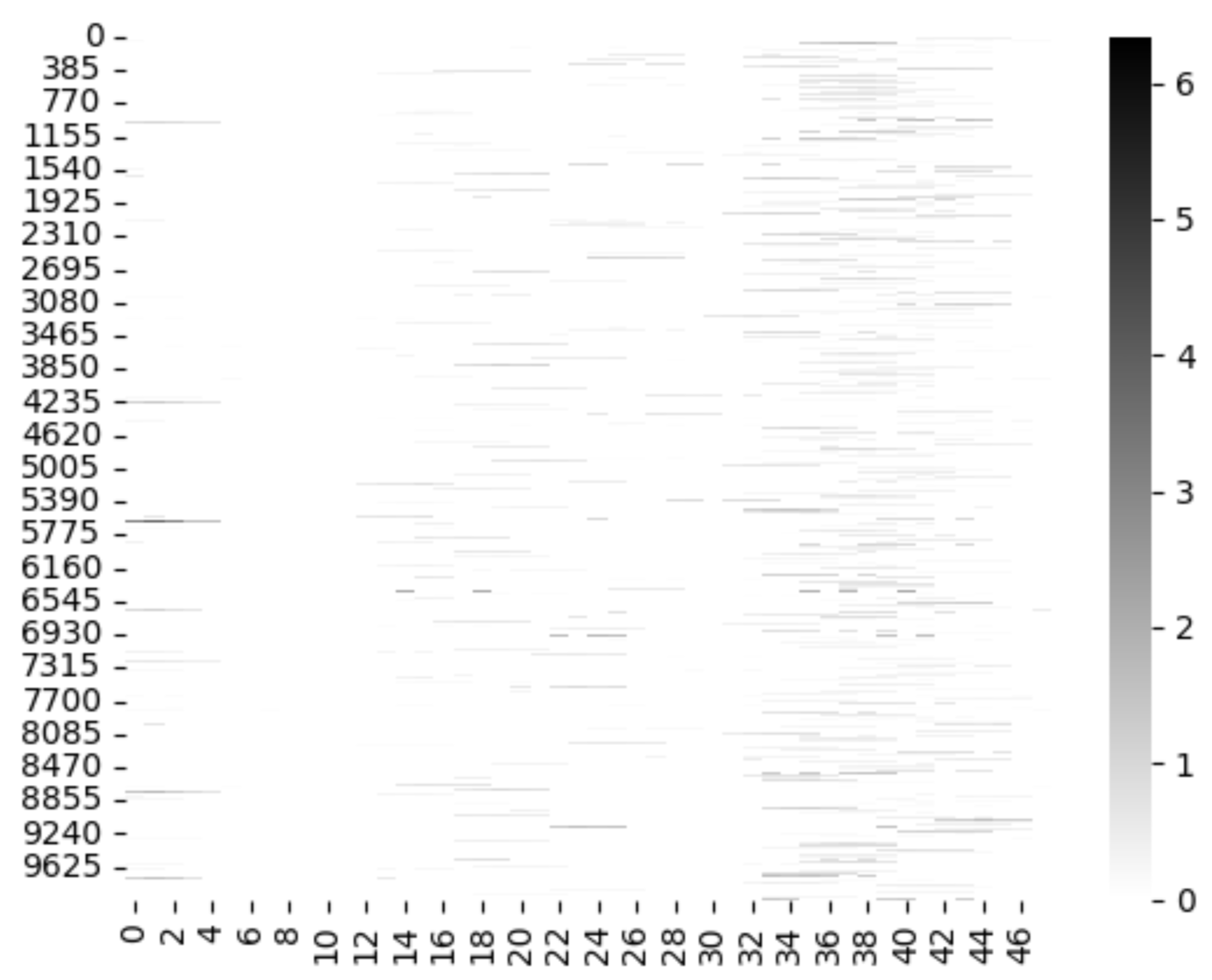}
    \caption{Heat map showing magnitude (in kWh) and position (in time) of peaks in the dataset. The x-axis shows the half hour settlement period throughout the day. Each y-value represents a distinct LCL daily profile (y-axis label indicates profile ID).}
    \label{fig:fig_peaks}
\end{figure}

We test fidelity metrics 1-5 in section \ref{sec:fidelity} on Faraday outputs trained on LCL daily profiles, using the methodologies described below. 
\begin{enumerate}
    \item Distribution of Autocorrelation Function (ACF) coefficients:
    We compute the ACF using the Pyro python package\cite{pyro}. Each time-series is a 48-dimensional vector, and a batch $N$ time-series will produce an autocorrelation matrix of $N \times 48$ dimensions of ACF coefficients. MMD\cite{mmd-paper} is used to quantify the distance between the distribution of the coefficients in the synthetic and real datasets.
    \item Standard statistical and distance measures:
    Mean and quantile daily consumption profiles (kWh) are calculated for the synthetic and real datasets. The `distance' between the mean/quantile synthetic and real profile is calculated by taking the sum of the absolute differences at each half hour. This provides a measure of the “mean and quantile deviation sum”. MMD is used to quantify the distances in distribution between synthetic and real data.
    \item Distribution of and timing of peaks: 
    First, prominent time series peaks are identified by simply taking the top $N$ values of the time series vector, and masking the rest of the vector with zeros. This produces a sparse time-series of only $N$ smart meter readings. MMD is used to measure the distance between the distribution of the ‘peaks’ in the synthetic and real datasets. The resulting dataset after masking the dataset is illustrated in the heat map in Fig.\ref{fig:fig_peaks}.
    \item Distribution of clusters:
    In this work, the data is clustered into an 25 clusters (chosen arbitrarily) using Gaussian Mixture Model \cite{scikit-learn}. As the cluster distribution could be obtained directly from the model, KL-Divergence is used instead of MMD to calculate the difference between the two distributions.
    \item Similarity of distribution at aggregated levels:
    To measure fidelity at the aggregated level, we take the clusters created in (4) and sum smart meter profiles within each cluster to obtain the cluster total consumption profile. We then compare synthetic and real datasets by computing the MMD distance between the ACFs, peak distributions, as well the mean absolute error and root mean squared error of the cluster totals.
\end{enumerate}

In section \ref{sec:fidelity_privacy_tradeoff}, we further investigate the trade-off between fidelity and privacy in the dataset using DP-SGD by altering the target $\epsilon$. The impact of dataset size is further considered by comparing metrics measured from synthetic datasets of size 50k, 100k, 150k and 200k.

\subsection{Measuring Utility}
As there are many potential downstream applications of synthetic data, we focus on two ‘fundamental’ tasks for synthetic smart meter data and apply these tasks to Faraday outputs trained on the LCL dataset:

\begin{enumerate}
    \item TSTR Classification: Perform a classification task to predict the season (Winter/Spring vs Summer/Autumn) from the smart meter profiles. 
    \item TSTR Forecasting: Predict the 48$^{th}$ half-hour average and 95$^{th}$ quantile consumption based on the previous 47 half-hours. 
\end{enumerate}
For both experiments, LCL daily profiles were divided into a TSTR ‘train set’ (2012 and 2013 data) and ‘evaluation set’ (2014). Synthetic models are trained on the TSTR train set, and TSTR classification and forecasting models are trained on the outputs of these synthetic models. The TSTR evaluation models are then evaluated on the evaluation set.

The TSTR classification task is a binary classification task where load profiles are used to predict the seasons Winter/ Spring (December to May) and Summer/ Autumn (June to November). The classification model is a standard multilinear perceptron. Months of year are `seen labels' in Faraday model for training.

The motivation for the TSTR forecasting task is to test whether time dynamics are preserved in synthetic smart meter data: if this is true, we expect good performance in predicting future demand. Similar to classification tasks, forecast models are trained on synthetic data produced using 2012-2013 LCL consumption data, and evaluated on 2014 consumption. We forecast both the average and 95$^{th}$ quantile demand. This is because grid operators are primarily concerned with peak demand on the grid which is explicitly captured in the 95$^{th}$ quantile forecasting task.

\section{Results}
\label{sec:results}

\subsection{Distance-based Reconstruction Attack}
\label{sec:distanced_based_reconstruction_attack_results}
We report the p-values as follows: the generative model trained without differential privacy is $0.999$ and the model trained with differential privacy $\epsilon=0.1$ is $0.397$. Both p-values are higher than 0.05, meaning we fail to reject the null hypothesis that synthetic data is closer than to the holdout data than to the training data. In other words, our test suggests that synthetic data is not closer to the training data than to the holdout data, indicating that memorisation has not occurred.

We note however that two-sample KS test works by comparing the cumulative distribution function (CDF) of two distributions, and is therefore more sensitive at the centre of the distribution and less in the tails (outlying region). As outliers are more prone to being memorised, a positive result from KS two-sample test does not offer the same guarantee to outliers. We need to also assess the privacy risk of outliers explicitly as described in \ref{sec:reconstruction_attack_method}.

\subsection{Outlier-Poisoned Reconstruction Attack}
\label{sec:reconstruction_attack_results}

\begin{figure*}
    \centering
    \includegraphics[width=0.9\linewidth]{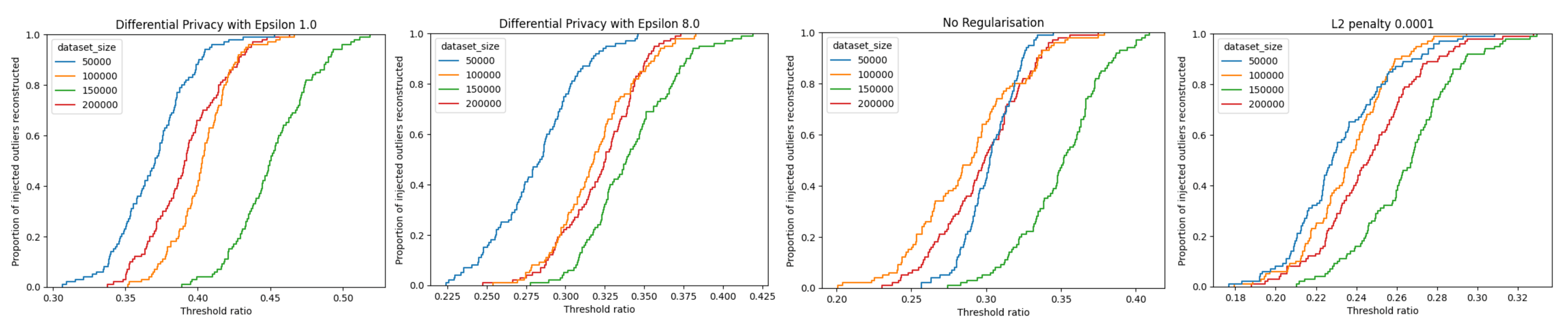}
    \caption{Effects of different privacy protection and dataset size on reconstruction attack. The success of the reconstruction attack is measured as the proportion of reconstructed outliers (y-axis). The x-axis shows varying threshold radius. }
    \label{fig:reconstruction_attack_results}
\end{figure*}

Fig.~\ref{fig:reconstruction_attack_results} shows the proportion of reconstructed outliers versus the threshold ratio for different dataset sizes with and without DP. The further the curve is to the right, the stronger the protection from reconstruction attacks. Across all dataset sizes, it is clear that DP with $\epsilon=1.0$ offers the best protection for privacy. Setting a threshold ratio of 0.3 with DP($\epsilon=1.0$) would lead to no outliers being reconstructed at all, whereas models trained with an L2 penalty of 0.0001 show virtually all outliers being reconstructed at the same threshold ratio. Moreover, results also show that $\epsilon=8.0$ offers little protection against reconstruction attack over no DP at all. 

The impact of dataset size on reconstruction attacks is inconclusive. Whilst the model trained with 50k dataset size has the weakest protection, models trained with dataset size of 200k dataset actually had less protection against reconstruction attack than the model trained with dataset size of 150k.

\subsection{Membership Inference Attack}

\label{sec:mia_experiment_results}

\begin{figure}
    \centering
    \includegraphics[width=\linewidth]{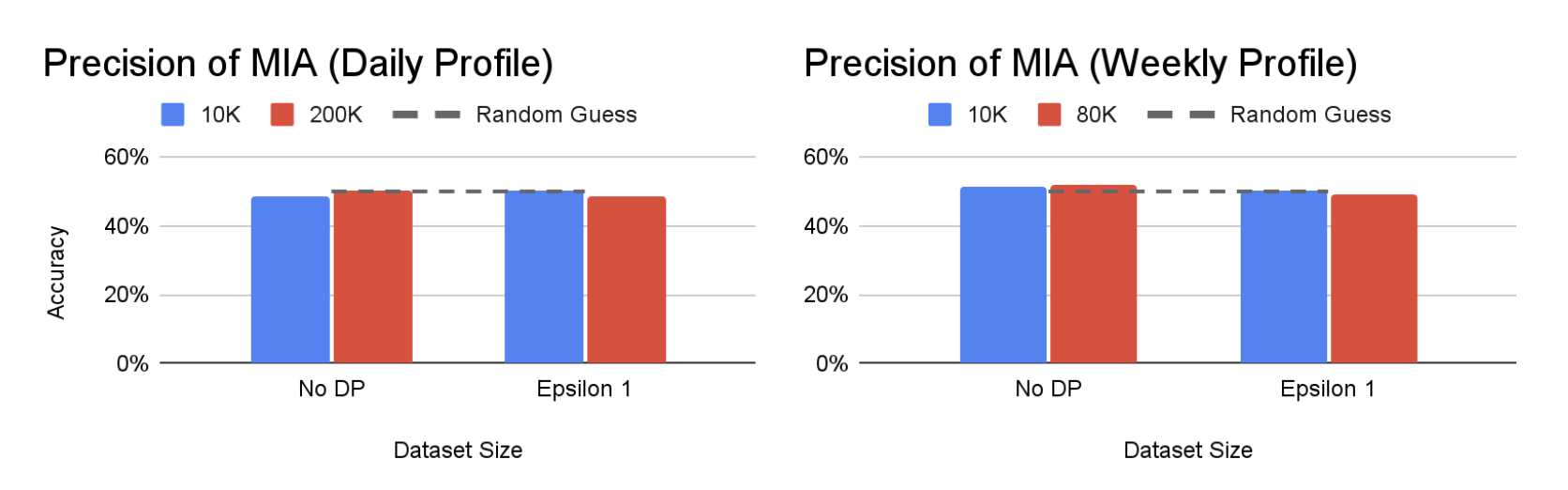}
    \caption{Precision of Membership Inference Attack.}
    \label{fig:mia_precision}
\end{figure}

In Fig \ref{fig:mia_precision}, we find that the MIA does not do a better job compared to a `random guess' at identifying whether a sample belongs to the training set (precision at $\sim50\%$ for all models). This however does not mean that synthetic smart meter data is inherently `private’. In cases of outliers, coupled with auxiliary datasets, it could still be possible to encounter privacy leaks. 

In particular, MIA is based on the premise that the training set is sufficiently different to the holdout set: by definition, the discriminator would fail if they are too similar, making MIA impossible and not suitable to identify overfitting/memorisation of the training data, especially around outlier values. Therefore, when training and holdout distributions are too similar, MIA will give a false sense of security. 

The results here are consistent with section \ref{sec:distanced_based_reconstruction_attack_results}. Standard privacy attack methods offer an optimistic view of privacy measures when the training data is too similar to holdout data. Whilst positive results offer guarantees on a population level, the same guarantees cannot be extended to outlying regions of the distribution. Given that MIA in its traditional form is likely insufficient to evaluate the privacy of smart meter data, due to training and holdout datasets potentially being too similar, we establish an `outlier-poisoned MIA'.

\subsection{Outlier-Poisoned MIA}
\label{sec:alternative_mia_experiment}

\begin{figure}
    \centering
    \includegraphics[width=0.9\linewidth]{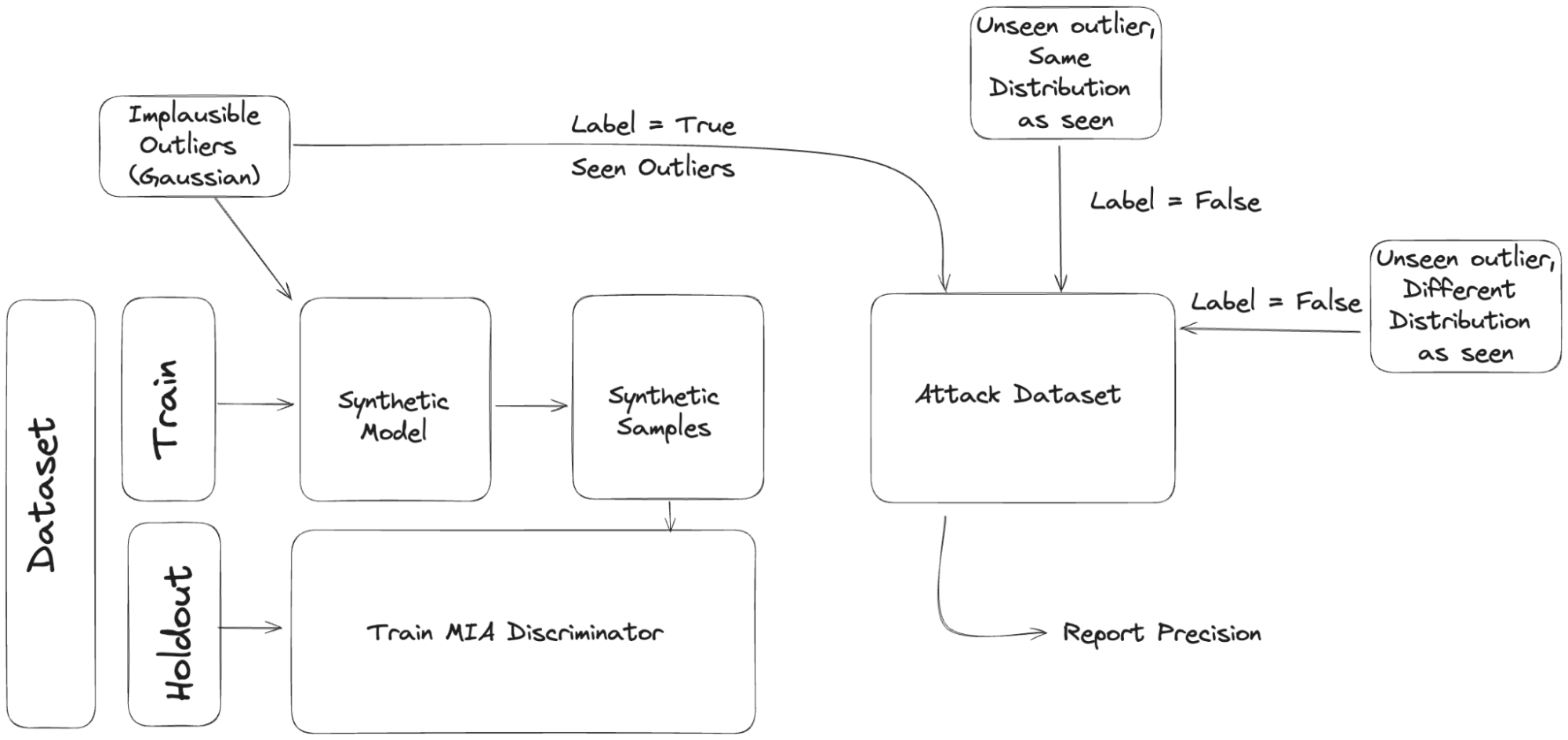}
    \caption{Training and attack protocol of membership inference attack (MIA) with poisoned dataset.}
    \label{fig:mia_poisoning_method}
\end{figure}

The same 100 implausible outliers used in Section \ref{sec:reconstruction_attack_method} are injected into the LCL training dataset. As before, synthetic samples are generated and together with a holdout sample and are used to train the MIA discriminator. MIA predicts whether suspected outliers were seen during training. If the synthetic model had memorised the training set, including the implausible outliers, they would appear in the synthetic samples and will be detected by the MIA discriminator.

An attack dataset is created comprising: 1) the 100 injected outliers seen during training (labelled true), 2) 100 unseen outliers drawn from the same distribution of the injected outliers (labelled false) and 3) 100 unseen outliers from a completely different distribution (labelled false). The size of the attack dataset is therefore 300 samples total. A random guess would result in precision of 0.33. Precision here means how many of the outliers that were predicted to be in the training set were actually in the training set. We explore the impact of varying the size of the synthetic dataset on the results, where we produce datasets of 50k, 100k, 150k and 200k (keeping the number of outliers constant). This method is outlined in Fig.~\ref{fig:mia_poisoning_method}.

We test the impact of varying $\epsilon$ in DP, and whether adding L2 penalty and dropout layers could mitigate privacy issues.

\begin{figure}
    \centering
    \includegraphics[width=\linewidth]{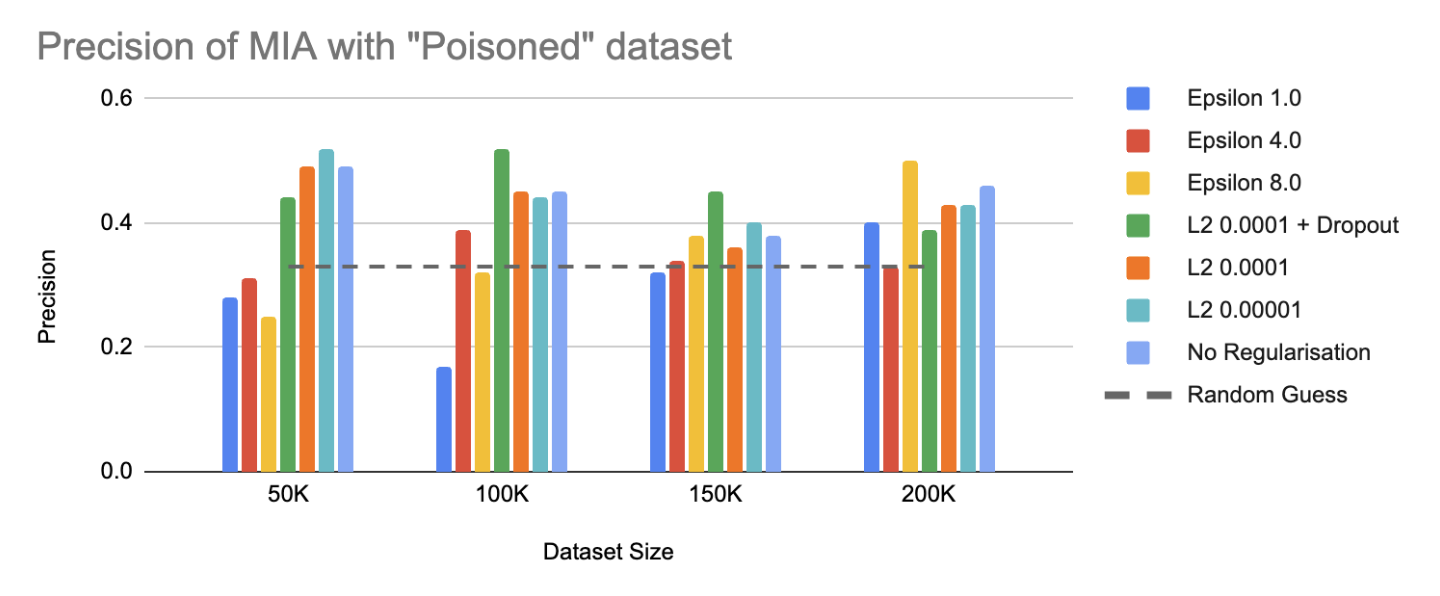}
    \caption{MIA precision metrics of poisoned dataset.}
    \label{fig:mia_poisoning_precision}
\end{figure}

Fig.~\ref{fig:mia_poisoning_precision} shows the precision results from MIA with outlier poisoning. Recall that the attack dataset comprises of 33\% true outliers seen during training. We therefore sort the predictions by probability and assigned the top 33rd percentile as `True', in line with the methodology described by Hayes et al,\cite{hayes2017logan}. The effect of this is such that the precision will be equal to the recall and thus F1-score, removing the effect of sensitivity of the discriminator's performance to its decision threshold.

Our results show that DP is effective at protecting against MIA in most settings, and is especially better than regularisation techniques at small dataset size. 

The ‘standard’ MIA approach is ineffective for testing privacy on smart meter datasets where training and holdout sets are similar. Injecting outliers during training and performing MIA specifically on these outliers allows us to more robustly detect privacy leaks. If these implausible outliers are protected or cannot be detected, then we can reason that all data points in the dataset are also likely protected.

\begin{figure*}[t!]
    \centering
    \includegraphics[width=\linewidth, height=6cm]{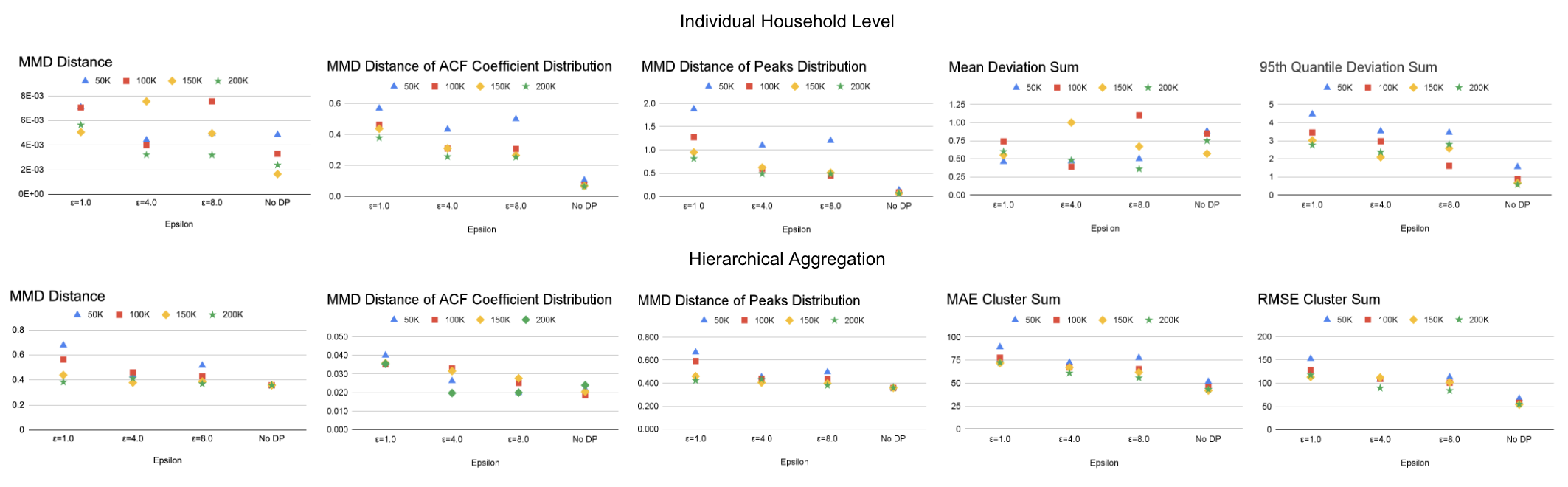}
    \caption{Fidelity metrics for different dataset sizes and levels of DP}
    \label{fig:fidelity_privacy_tradeoff}
\end{figure*}

\subsection{Fidelity versus Privacy Trade-off}
\label{sec:fidelity_privacy_tradeoff}

In Fig.~\ref{fig:fidelity_privacy_tradeoff} we present the results fidelity metrics defined in Section \ref{sec:fidelity} and how they vary with varying levels of privacy protection and dataset size. In general, we find higher privacy protection (lower $\epsilon$) results in larger distances (lower fidelity) between the distributions of real and synthetic datasets. Furthermore, for the same epsilon, a larger dataset reduces the distances  (higher fidelity) between the distributions.

This result approximately holds for the quantile deviation sum, however, clearly it does not hold for the mean deviation sum where there is no trend with $\epsilon$. This could be an indication that the generative model captured the 95$^{th}$ quantile well, but is struggling to represent the mean well, and would benefit from more optimisation e.g. hyper-parameter tuning.

\subsection{Utility versus Privacy Trade-off}
\label{sec:utility_privacy_tradeoff}


\begin{figure}
    \centering
    \includegraphics[width=\linewidth]{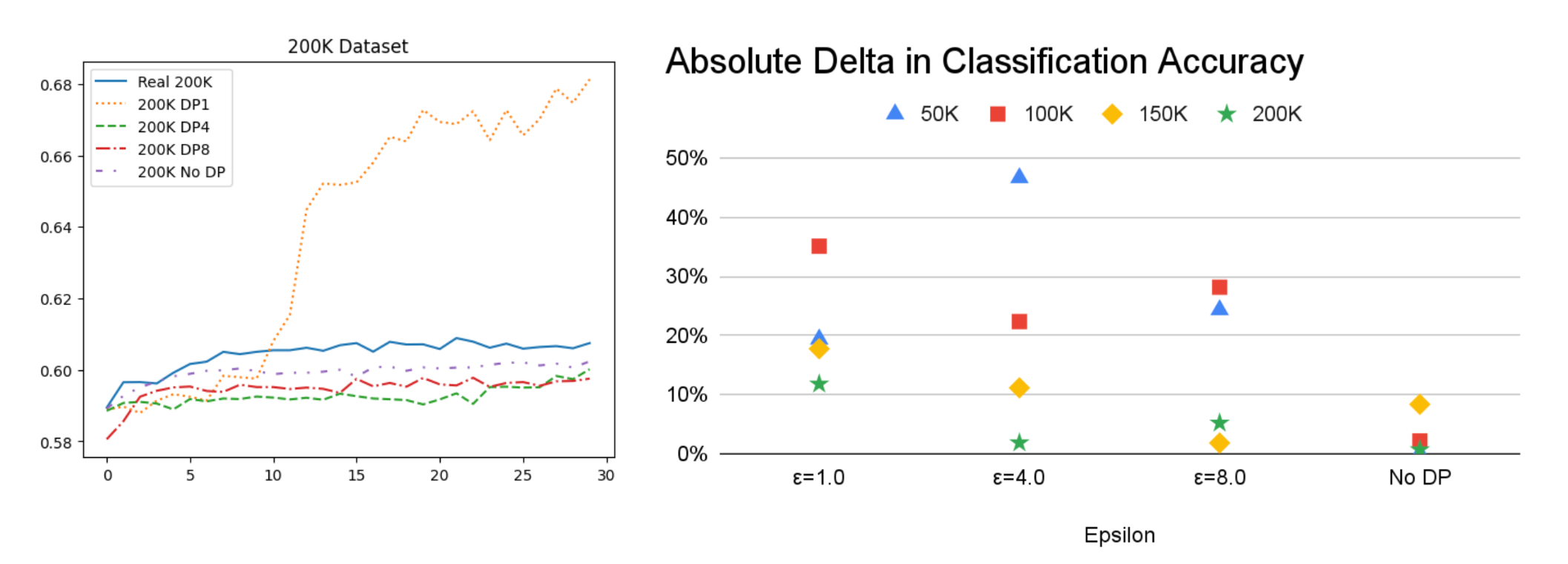}
    \caption{Left: Classification accuracy vs epoch. Right: Absolute difference in test accuracy of real vs synthetic data.}
    \label{fig:tstr_class_accuracy}
\end{figure}

As shown in Fig \ref{fig:tstr_class_accuracy}, we find that the TSTR classification model trained on on synthetic data without DP showed similar accuracy to the model trained real data. Increasing the level of privacy protection results in a classification accuracy offset from the non-DP models which is especially true for DP$(\epsilon = 1.0)$ which shows a significantly higher classification accuracy. These results are consistent for datasets of varying sizes (tested with N = 50K, 100K, 150K, 200K). 

In section \ref{sec:fidelity_privacy_tradeoff} we show that DP$(\epsilon = 1.0)$ models often result in  lower fidelity compared to no DP models. This can be seen when looking at the distribution of peaks in smart meter data (Fig \ref{fig:fidelity_privacy_tradeoff}), which could potentially lead to seasonality being easier to predict using the synthetic dataset compared to the real dataset, i.e., resulting in higher accuracy when conducting the utility prediction task. 

However, we point out that achieving higher accuracy in a utility task on synthetic data (compared to real data) does not equate to higher utility of the synthetic data model. For example, DP$(\epsilon = 1.0)$ could in principle lead to synthetic data that is less diverse by mitigating `outliers' during training (see section \ref{sec:reconstruction_attack_results}), resulting in an improved classification model compared to `messier' raw data (without DP). For synthetic data to be a faithful representation of the real data, we require the difference in performance to be similar, not worse or better. We should therefore compare the absolute difference in the accuracy of TSTR classifiers trained on synthetic vs real data as a measure of utility.

Fig \ref{fig:tstr_class_accuracy} shows that models with no DP have very similar performance compared to models trained with real data. DP leads to significant differences in classification accuracy, but that can be somewhat mitigated with a larger dataset size.


\begin{figure}
    \centering
    \includegraphics[width=\linewidth]{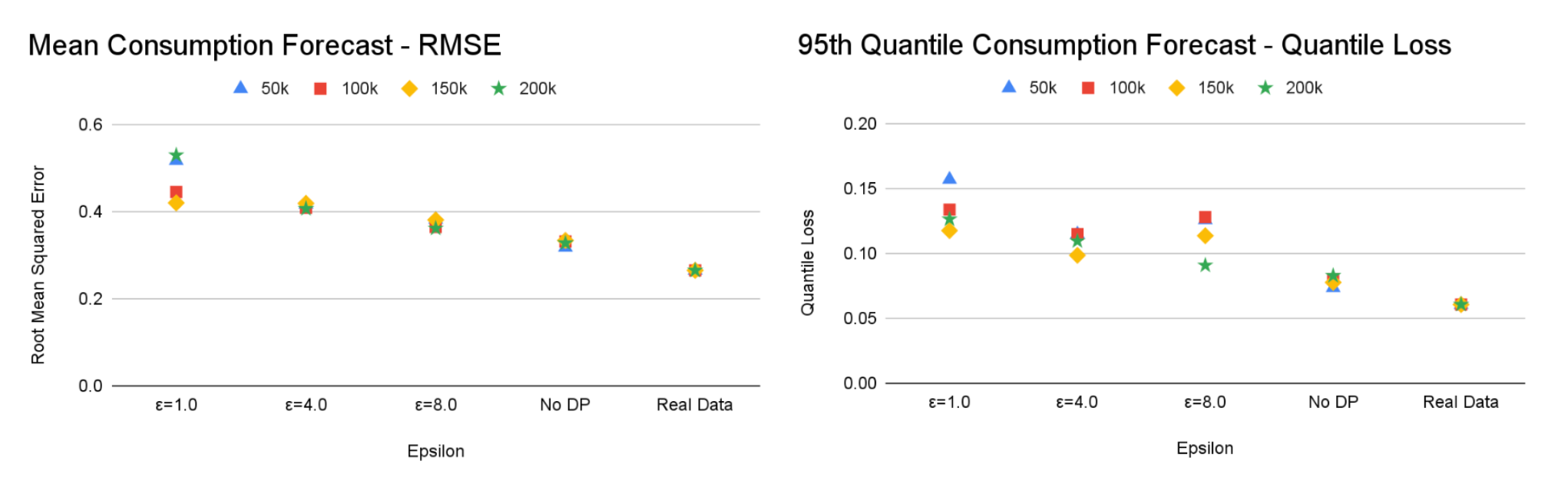}
    \caption{Root mean squared error for average consumption (left) and Quantile loss for 95$^{th}$ quantile consumption (right).}
    \label{fig:utility_forecasting_results}
\end{figure}

The results for the TSTR forecasting are shown in Figure \ref{fig:utility_forecasting_results}. For both forecasting tasks, we find the forecasting performance to worsen with increasing privacy protection (lower $\epsilon$). Whilst increasing dataset size failed to improve the accuracy of the mean forecast, there's some evidence to suggest that dataset size can improve quantile regression performance. This is consistent with what we saw in \ref{sec:fidelity_privacy_tradeoff}.

\section{Discussion}
\label{sec:summaryofresults}

In this section, we comment on results of all sections and use this exercise as a case study of how practitioners could use the fidelity, utility and privacy framework to quantify the quality of synthetic smart meter data.

\subsection{Fidelity}

Fidelity metrics allow us to identify which characteristics of the smart meter data that the algorithm could or could not preserve and identify potential mitigation tactics. In this example, we find that $\epsilon=1.0$ severely compromises the fidelity of the results, but there is evidence that increasing dataset size could improve fidelity for a given $\epsilon$, except for the dataset mean (mean deviation sum metric). This could imply areas for improvement in the algorithm on LCL dataset that we need to investigate further (e.g. better hyper-parameter tuning or architectural tweaks) to better capture the mean.

\subsection{Utility}
Fidelity metrics only allow us to identify where the model is failing. Utility metrics help us identify how much inaccuracy we can tolerate whilst having the data remain useful enough for real-world tasks. On classification tasks, we find that whilst a 200K dataset with $\epsilon=1.0$ leads to significant deterioration of results on classification, mean forecast and 95$^{th}$ quantile forecast accuracy, it could still support $\epsilon=4.0$.


\subsection{Privacy}

Given that the performance of the reconstruction attack depends on the $\epsilon$, we propose that for all models and applications, we always perform explicit privacy analysis, instead of relying implicitly on DP or dataset size to make any privacy claims. This however comes at a significant computational cost (DP models on average take about 10 times as long to train compared to non-DP models). Increasing dataset size however does not appear to affect privacy directly.

We find that from MIA results, normal regularisation techniques like L2 penalty or dropouts offer no privacy protection - differential privacy is the only mechanism that reduces the effects of an MIA attack to be similar to that of random guess. 

Stakeholders could choose a threshold radius for reconstruction attack that suits their risk-appetite and application. For the application where synthetic data would be shared publicly for perpetuity, we propose a conservative threshold of 0.3. This means for example that for an outlier with daily total consumption of 30kWh, a 0.3 threshold means $30\pm{9}$ kWh would compromise the outlier. Using 0.3 as our threshold, we find that $\epsilon=8.0$ would compromise about 25$\%$ of the dataset whilst $\epsilon=1.0$ offers 100$\%$ protection. In fact, $\epsilon=8.0$ showed little to no improvement over the model trained without DP. 

\subsection{Appropriate Distance Metric for Privacy}
\label{sec:privacy_radius}

\begin{figure}
    \centering
    \includegraphics[width=0.8\linewidth]{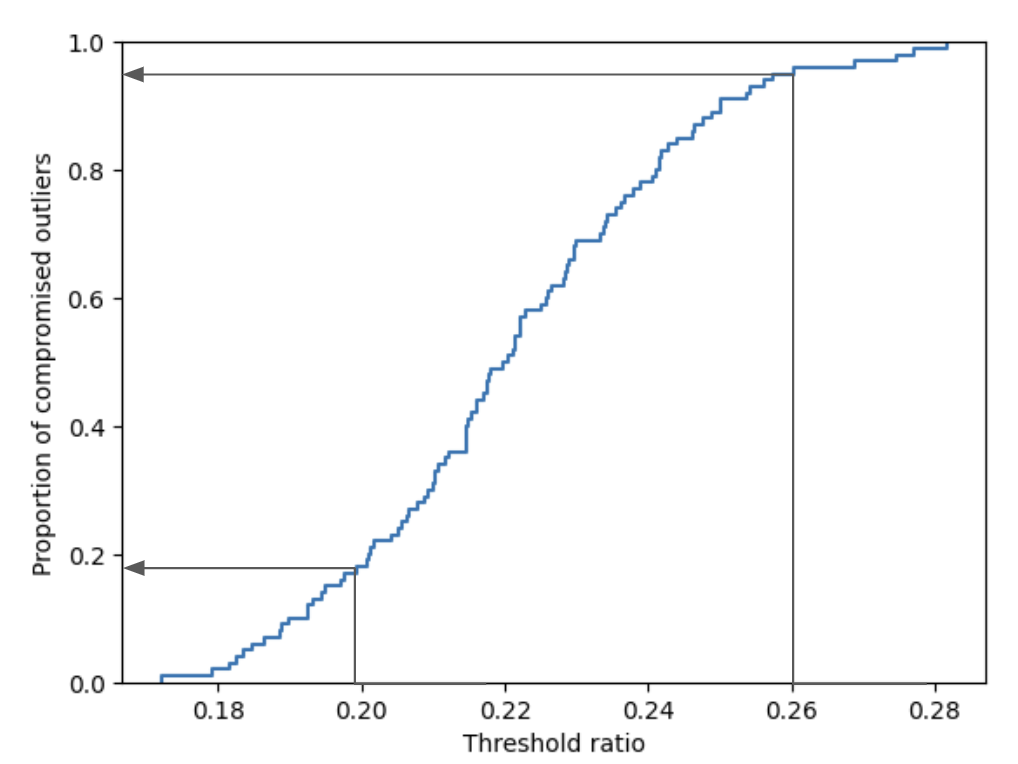}
    \caption{Cumulative distribution function (CDF) showing the \% of outliers  considered to be reconstructed based as a function of the threshold ratio $r$.
}
    \label{fig:cummulative_frac}
\end{figure}
In Section \ref{sec:reconstruction_attack_method} we outlined a novel method of measuring the threshold radius for an data point to be considered reconstructed. By setting the threshold radius as a ratio to the vector’s norm, we take into account the size of the vector itself, allowing stakeholders to develop a risk policy that’s consistent across different settings by defining the value of $r$. An outlier with small magnitude would have a small radius, whilst an outlier with larger magnitude would have a larger radius as a result. Defining the threshold radius in this way is essentially like having a percentage error around each injected outlier, where if synthetic data falls within the error bounds it is considered `too close' and thus considered a privacy leak.

In Fig \ref{fig:cummulative_frac}, we show the proportion of outliers which are reconstructed as a function of $r$. In this example, setting a $r = 0.2$ shows that around 18\% of artificial outliers have a generated sample being too close to it, whilst a threshold $r = 0.26$ shows that 95\% of artificial outliers have at least 1 generated sample being too close. Risky applications might require larger distance radius, whilst less risky applications could afford tighter radius. This radius is referred to as the `privacy threshold' for the reconstruction attack. Practitioners can compare different models (e.g. with different $\epsilon$ values) to quantify impact on reconstruction attacks. The more the CDF in Fig \ref{fig:cummulative_frac} is to the right, the stronger the privacy protection.

\subsection{Impact of Dataset Size on Fidelity, Utility and Privacy}

In this work, we have found that dataset size has no direct effect on privacy. Intuitively this would make sense given the nature of how privacy accounting mechanism in DP-SGD algorithm works. When we train an ML algorithm using DP-SGD with a target $\epsilon$ value, just enough noise is added during gradient descent to result achieve that $\epsilon$ regardless of the size of training data. Dataset size however is still important as we have shown that for a given $\epsilon$ value, larger dataset size can achieve higher fidelity and utility metrics.

\subsection{Communicating privacy to stakeholders}
Communicating the real-world impact of $\epsilon$ to stakeholders is vital given that the privacy of sensitive data is at stake. $\epsilon$ is not an not an absolute measure of privacy risk; $\epsilon$ is the marginal increase in probability or risk that private data is leaked. What this means for the protection of their sensitive data however is not obvious. There is also a lack of consensus amongst practitioners as to what is an acceptable range of $\epsilon$ values. DP should be accompanied alongside explicit tests for privacy to translate $\epsilon$ into something more meaningful for stakeholders.

\section{Conclusion}
\label{sec:conclusion}
In this work, we take commonly used concepts for evaluating synthetic data: fidelity, utility and privacy, and investigate their application to synthetic smart meter data. Evaluating privacy is essential when generating synthetic smart meter data given the sensitive nature of the training data for the generative model. Although DP-SGD is a state-of-art method for privacy protection, an appropriate value for $\epsilon$ is subjective. Furthermore there is a challenge in explaining to stakeholders what this actually means for the protection of sensitive smart meter data which is protected under GDPR. Privacy attacks are necessary to determine risks of private data being leaked. Their results are also more interpretable than the notion of $\epsilon$ from differential privacy.

We also show that `standard' privacy attacks methods on smart meter datasets give overly optimistic results and cannot quantify the privacy risks to outliers. We propose an additional method of launching privacy attacks by training generative models with implausible outliers injected into the training set, and launching privacy attacks directly on outliers seen during training. We also compared DP-SGD against common regularisation techniques and showed that DP-SGD consistently outperforms regularisation techniques at protecting privacy. 

Although we find dataset size has no direct impact on privacy protection, increasing dataset size significantly increases the fidelity and utility synthetic data for a specific value of $\epsilon$. Privacy and fidelity/utility trade-off is an iterative process. Practitioners should start off by declaring the privacy requirement for their application. Based on this, they identify the $\epsilon$ required, and look at the impact on fidelity and utility. If the practitioner finds that the current dataset size is insufficient to satisfy both privacy and fidelity/utility requirements, the practitioner should either procure more data or if procuring more data is infeasible or too expensive, conclude that it is impossible to produce a synthetic dataset that is private whilst remaining useful. Practitioners should not be relaxing $\epsilon$ without understanding that doing so compromises on privacy protection, and cannot simply rely on the fact that they have implemented differential privacy to make privacy protection claims (e.g., as in the case of the LCL dataset, where we saw that $\epsilon=8.0$ was insufficient). 

Applications of synthetic smart meter data are highly valuable to accelerate the Net Zero transition. Potential use cases include demand models in future energy scenarios or the development of next-generation optimization and control techniques to enable the management of decarbonized grids, for example. In order for synthetic outputs to be useful for their intended purposes, it is important that key characteristics of the data are preserved. In Sections \ref{sec:fidelity} and \ref{sec:utility}, we recommend a suite of fidelity and utility metrics specific to evaluating the quality of smart meter data and the preservation of these characteristics.

\subsection{Future Work}

This paper proposes a common evaluation framework to benchmark algorithms to generate synthetic smart meter data, drawing inspiration from work already done in other areas e.g. health and finance. 

With regards to fidelity and utility, we have only proposed a limited suite of metrics and evaluation tasks that focuses on foundational characteristics and use cases. There could be other tasks that users of synthetic smart meter data require that are also worth taking into account, such as demand ramping.

Furthermore, most generative models address fidelity (through distance- or accuracy-based loss functions e.g. MMD, KL-divergence etc.) and privacy (through differential privacy) explicitly in the training process, but the training process generally does not address utility. Another direction of future work might include explicitly incorporating considerations of utility within the training or fine-tuning processes for synthetic data generation models (just as is done for fidelity and privacy metrics). For instance, prior work in the `predict-then-optimize' setting has proposed training forecasting models in a way that is maximally useful for downstream optimization tasks\cite{donti2017task},\cite{kotary_end_to_end_2021},\cite{elmachtoub_smart_2022}; similarly, synthetic data generation models could be trained or fine-tuned on objectives that explicitly capture the utility of the generated data for downstream tasks.

On the privacy front, whilst we have looked into reconstruction and membership inference attacks, we were not able to conduct analyses on attribute inference attacks~\cite{jordon2022}. The Low Carbon London dataset only has data on household IDs and their consumption; it does not have any data on sensitive household characteristics such as their demographics to allow us to perform attribute inference attacks explicitly. Therefore, we could only quantify the risk of data being leaked and cannot quantify the damage on individuals should such a leak occur. We welcome researchers with access to both smart meter data and household characteristics to research attribute inference risks and potentially open-source models that can help quantify the risk of sensitive information being revealed in synthetic smart meter data.

To date, granular smart meter data remains largely inaccessible. Privacy legislation, security controls and technical barriers may continue to limit the ability of networks or suppliers to share real data. Synthetic data has the potential to overcome these challenges. However, further work is needed to study attribute inference attacks to quantify the risks of sensitive information being revealed in synthetic smart meter data. In addition, we recommend that governments and regulatory bodies consider how synthetic data can be used while addressing privacy and other ethical concerns, either through regulations or appropriate licensing. The energy industry should also think about safeguards to limit the possibility of synthetic smart meter data from being misused, similar to existing regulation and laws in the financial services industry in the UK\cite{fca} and the US\cite{civil_rights_div}.

\bibliographystyle{IEEEtran}
\bibliography{main}

\begin{thebibliography}{10}
\providecommand{\url}[1]{#1}
\csname url@samestyle\endcsname
\providecommand{\newblock}{\relax}
\providecommand{\bibinfo}[2]{#2}
\providecommand{\BIBentrySTDinterwordspacing}{\spaceskip=0pt\relax}
\providecommand{\BIBentryALTinterwordstretchfactor}{4}
\providecommand{\BIBentryALTinterwordspacing}{\spaceskip=\fontdimen2\font plus
\BIBentryALTinterwordstretchfactor\fontdimen3\font minus \fontdimen4\font\relax}
\providecommand{\BIBforeignlanguage}[2]{{%
\expandafter\ifx\csname l@#1\endcsname\relax
\typeout{** WARNING: IEEEtran.bst: No hyphenation pattern has been}%
\typeout{** loaded for the language `#1'. Using the pattern for}%
\typeout{** the default language instead.}%
\else
\language=\csname l@#1\endcsname
\fi
#2}}
\providecommand{\BIBdecl}{\relax}
\BIBdecl

\bibitem{esc_open_sm_data_call}
\BIBentryALTinterwordspacing
E.~S. Catapult, ``{Data for Good, Smart Meter Access},'' Energy Systems Catapult, Tech. Rep., 10 2023. [Online]. Available: \url{https://es.catapult.org.uk/report/data-for-good-smart-meter-data-access/}
\BIBentrySTDinterwordspacing

\bibitem{beckel2014revealing}
C.~Beckel, L.~Sadamori, T.~Staake, and S.~Santini, ``Revealing household characteristics from smart meter data,'' \emph{Energy}, vol.~78, pp. 397--410, 2014.

\bibitem{radovanovic2022unique}
D.~Radovanovic, A.~Unterweger, G.~Eibl, D.~Engel, and J.~Reichl, ``How unique is weekly smart meter data?'' \emph{Energy Informatics}, vol.~5, no. Suppl 1, p.~13, 2022.

\bibitem{jordon2022}
J.~Jordon, L.~Szpruch, F.~Houssiau, M.~Bottarelli, G.~Cherubin, C.~Maple, S.~N. Cohen, and A.~Weller, ``Synthetic data--what, why and how?'' \emph{arXiv preprint arXiv:2205.03257}, 2022.

\bibitem{haben2021review}
S.~Haben, S.~Arora, G.~Giasemidis, M.~Voss, and D.~V. Greetham, ``Review of low voltage load forecasting: Methods, applications, and recommendations,'' \emph{Applied Energy}, vol. 304, p. 117798, 2021.

\bibitem{pang2018hierarchical}
Y.~Pang, B.~Yao, X.~Zhou, Y.~Zhang, Y.~Xu, and Z.~Tan, ``Hierarchical electricity time series forecasting for integrating consumption patterns analysis and aggregation consistency.'' in \emph{IJCAI}, 2018, pp. 3506--3512.

\bibitem{kingma_vae}
\BIBentryALTinterwordspacing
D.~P. Kingma and M.~Welling, ``Auto-{Encoding} {Variational} {Bayes},'' Dec. 2022, arXiv:1312.6114 [cs, stat]. [Online]. Available: \url{http://arxiv.org/abs/1312.6114}
\BIBentrySTDinterwordspacing

\bibitem{goodfellow_generative_2014}
\BIBentryALTinterwordspacing
I.~J. Goodfellow, J.~Pouget-Abadie, M.~Mirza, B.~Xu, D.~Warde-Farley, S.~Ozair, A.~Courville, and Y.~Bengio, ``Generative {Adversarial} {Networks},'' Jun. 2014, arXiv:1406.2661 [cs, stat]. [Online]. Available: \url{http://arxiv.org/abs/1406.2661}
\BIBentrySTDinterwordspacing

\bibitem{lin_diffusion_2023}
\BIBentryALTinterwordspacing
L.~Lin, Z.~Li, R.~Li, X.~Li, and J.~Gao, ``Diffusion {Models} for {Time} {Series} {Applications}: {A} {Survey},'' Apr. 2023, arXiv:2305.00624 [cs]. [Online]. Available: \url{http://arxiv.org/abs/2305.00624}
\BIBentrySTDinterwordspacing

\bibitem{netflix-anonymity}
A.~Narayanan and V.~Shmatikov, ``How to break anonymity of the netflix prize dataset (2006),'' \emph{arXiv preprint cs/0610105}, 2016.

\bibitem{sweeney2000simple}
L.~Sweeney, ``Simple demographics often identify people uniquely,'' \emph{Health (San Francisco)}, vol. 671, no. 2000, pp. 1--34, 2000.

\bibitem{faraday_paper}
\BIBentryALTinterwordspacing
S.~Chai and G.~Chadney, ``Faraday: {Synthetic} {Smart} {Meter} {Generator} for the smart grid,'' 2024. [Online]. Available: \url{http://arxiv.org/abs/2404.04314}
\BIBentrySTDinterwordspacing

\bibitem{low_carbon_london}
{UK Power Networks}, ``Smartmeter energy consumption data in london households,'' 2014, smart meter data retrieved from London Datastore, \url{https://data.london.gov.uk/dataset/smartmeter-energy-use-data-in-london-households}.

\bibitem{dwork2006calibrating}
C.~Dwork, F.~McSherry, K.~Nissim, and A.~Smith, ``Calibrating noise to sensitivity in private data analysis,'' in \emph{Theory of Cryptography: Third Theory of Cryptography Conference, TCC 2006, New York, NY, USA, March 4-7, 2006. Proceedings 3}.\hskip 1em plus 0.5em minus 0.4em\relax Springer, 2006, pp. 265--284.

\bibitem{dwork2006our}
C.~Dwork, K.~Kenthapadi, F.~McSherry, I.~Mironov, and M.~Naor, ``Our data, ourselves: Privacy via distributed noise generation,'' in \emph{Advances in Cryptology-EUROCRYPT 2006: 24th Annual International Conference on the Theory and Applications of Cryptographic Techniques, St. Petersburg, Russia, May 28-June 1, 2006. Proceedings 25}.\hskip 1em plus 0.5em minus 0.4em\relax Springer, 2006, pp. 486--503.

\bibitem{shahani2023techniques}
S.~Shahani, J.~Abraham, and Venkateswaran, ``Techniques for privacy-preserving data aggregation in an untrusted distributed environment,'' in \emph{Proceedings of the 6th Joint International Conference on Data Science \& Management of Data (10th ACM IKDD CODS and 28th COMAD)}, 2023, pp. 286--287.

\bibitem{real-world-dp-eps}
D.~Desfontaines, ``A list of real-world uses of differential privacy,'' \url{https://desfontain.es/blog/real-world-differential-privacy.html}, 10 2021, ted is writing things (personal blog).

\bibitem{dp-choosing-epsilon}
J.~Hsu, M.~Gaboardi, A.~Haeberlen, S.~Khanna, A.~Narayan, B.~C. Pierce, and A.~Roth, ``Differential privacy: An economic method for choosing epsilon,'' in \emph{2014 IEEE 27th Computer Security Foundations Symposium}.\hskip 1em plus 0.5em minus 0.4em\relax IEEE, 2014, pp. 398--410.

\bibitem{how-much-is-enough-dp}
J.~Lee and C.~Clifton, ``How much is enough? choosing $\varepsilon$ for differential privacy,'' in \emph{Information Security: 14th International Conference, ISC 2011, Xi’an, China, October 26-29, 2011. Proceedings 14}.\hskip 1em plus 0.5em minus 0.4em\relax Springer, 2011, pp. 325--340.

\bibitem{synthetic-data-privacy}
V.~Marshall, C.~Markham, P.~Avramovic, P.~Comerford, C.~Maple, and L.~Szpruch, ``Exploring synthetic data validation – privacy, utility and fidelity,'' 2024, \url{https://www.fca.org.uk/publications/research-articles/exploring-synthetic-data-validation-privacy-utility-fidelity} [Accessed: 2024-05-24.

\bibitem{deep-learning-w-dp}
M.~Abadi, A.~Chu, I.~Goodfellow, H.~B. McMahan, I.~Mironov, K.~Talwar, and L.~Zhang, ``Deep learning with differential privacy,'' in \emph{Proceedings of the 2016 ACM SIGSAC conference on computer and communications security}, 2016, pp. 308--318.

\bibitem{gdt-descent}
P.~Baldi, ``Gradient descent learning algorithm overview: A general dynamical systems perspective,'' \emph{IEEE Transactions on neural networks}, vol.~6, no.~1, pp. 182--195, 1995.

\bibitem{pytorch-opacus}
A.~Yousefpour, I.~Shilov, A.~Sablayrolles, D.~Testuggine, K.~Prasad, M.~Malek, J.~Nguyen, S.~Ghosh, A.~Bharadwaj, J.~Zhao \emph{et~al.}, ``Opacus: User-friendly differential privacy library in pytorch,'' \emph{arXiv preprint arXiv:2109.12298}, 2021.

\bibitem{esteban2017real}
C.~Esteban, S.~L. Hyland, and G.~R{\"a}tsch, ``Real-valued (medical) time series generation with recurrent conditional gans,'' \emph{arXiv preprint arXiv:1706.02633}, 2017.

\bibitem{pratt_kolmogorov-smirnov_1981}
\BIBentryALTinterwordspacing
J.~W. Pratt and J.~D. Gibbons, ``\BIBforeignlanguage{en}{Kolmogorov-{Smirnov} {Two}-{Sample} {Tests}},'' in \emph{\BIBforeignlanguage{en}{Concepts of {Nonparametric} {Theory}}}, J.~W. Pratt and J.~D. Gibbons, Eds.\hskip 1em plus 0.5em minus 0.4em\relax New York, NY: Springer, 1981, pp. 318--344. [Online]. Available: \url{https://doi.org/10.1007/978-1-4612-5931-2_7}
\BIBentrySTDinterwordspacing

\bibitem{bounliphone_test_2016}
\BIBentryALTinterwordspacing
W.~Bounliphone, E.~Belilovsky, M.~B. Blaschko, I.~Antonoglou, and A.~Gretton, ``A {Test} of {Relative} {Similarity} {For} {Model} {Selection} in {Generative} {Models},'' Feb. 2016, arXiv:1511.04581 [cs, stat]. [Online]. Available: \url{http://arxiv.org/abs/1511.04581}
\BIBentrySTDinterwordspacing

\bibitem{parikh2022canary}
R.~Parikh, C.~Dupuy, and R.~Gupta, ``Canary extraction in natural language understanding models,'' \emph{arXiv preprint arXiv:2203.13920}, 2022.

\bibitem{hayes2017logan}
J.~Hayes, L.~Melis, G.~Danezis, and E.~De~Cristofaro, ``Logan: Membership inference attacks against generative models,'' \emph{arXiv preprint arXiv:1705.07663}, 2017.

\bibitem{shokri2017membership}
R.~Shokri, M.~Stronati, C.~Song, and V.~Shmatikov, ``Membership inference attacks against machine learning models,'' in \emph{2017 IEEE symposium on security and privacy (SP)}.\hskip 1em plus 0.5em minus 0.4em\relax IEEE, 2017, pp. 3--18.

\bibitem{tao2021benchmarking}
Y.~Tao, R.~McKenna, M.~Hay, A.~Machanavajjhala, and G.~Miklau, ``Benchmarking differentially private synthetic data generation algorithms,'' \emph{arXiv preprint arXiv:2112.09238}, 2021.

\bibitem{diversity-in-synth-ts}
F.~Bahrpeyma, M.~Roantree, P.~Cappellari, M.~Scriney, and A.~McCarren, ``A methodology for validating diversity in synthetic time series generation,'' \emph{MethodsX}, vol.~8, p. 101459, 2021.

\bibitem{mmd-paper}
A.~Gretton, K.~M. Borgwardt, M.~J. Rasch, B.~Sch{\"o}lkopf, and A.~Smola, ``A kernel two-sample test,'' \emph{The Journal of Machine Learning Research}, vol.~13, no.~1, pp. 723--773, 2012.

\bibitem{js-divergence}
F.~Nielsen, ``On the jensen--shannon symmetrization of distances relying on abstract means,'' \emph{Entropy}, vol.~21, no.~5, p. 485, 2019.

\bibitem{ts-gan}
J.~Yoon, D.~Jarrett, and M.~Van~der Schaar, ``Time-series generative adversarial networks,'' \emph{Advances in neural information processing systems}, vol.~32, 2019.

\bibitem{hndyman_forecasting}
\BIBentryALTinterwordspacing
R.~Hyndmand and G.~Athanasopoulos, \emph{Forecasting: principles and practice, 3rd edition}.\hskip 1em plus 0.5em minus 0.4em\relax Melbourne, Australia: OTexts, 2021. [Online]. Available: \url{https://otexts.com/fpp3/}
\BIBentrySTDinterwordspacing

\bibitem{tesauro_scaling_weightdecay}
G.~Tesauro and B.~Janssens, ``Scaling relationships in back-propagation learning,'' \emph{Complex Systems}, vol.~2, no.~1, pp. 39--44, 1988.

\bibitem{srivastava_dropout_2014}
\BIBentryALTinterwordspacing
N.~Srivastava, G.~Hinton, A.~Krizhevsky, I.~Sutskever, and R.~Salakhutdinov, ``Dropout: {A} {Simple} {Way} to {Prevent} {Neural} {Networks} from {Overfitting},'' \emph{Journal of Machine Learning Research}, vol.~15, no.~56, pp. 1929--1958, 2014. [Online]. Available: \url{http://jmlr.org/papers/v15/srivastava14a.html}
\BIBentrySTDinterwordspacing

\bibitem{pyro}
\BIBentryALTinterwordspacing
E.~Bingham, J.~P. Chen, M.~Jankowiak, F.~Obermeyer, N.~Pradhan, T.~Karaletsos, R.~Singh, P.~A. Szerlip, P.~Horsfall, and N.~D. Goodman, ``Pyro: Deep universal probabilistic programming,'' \emph{J. Mach. Learn. Res.}, vol.~20, pp. 28:1--28:6, 2019. [Online]. Available: \url{http://jmlr.org/papers/v20/18-403.html}
\BIBentrySTDinterwordspacing

\bibitem{scikit-learn}
F.~Pedregosa, G.~Varoquaux, A.~Gramfort, V.~Michel, B.~Thirion, O.~Grisel, M.~Blondel, P.~Prettenhofer, R.~Weiss, V.~Dubourg, J.~Vanderplas, A.~Passos, D.~Cournapeau, M.~Brucher, M.~Perrot, and E.~Duchesnay, ``Scikit-learn: Machine learning in {P}ython,'' \emph{Journal of Machine Learning Research}, vol.~12, pp. 2825--2830, 2011.

\bibitem{donti2017task}
P.~Donti, B.~Amos, and J.~Z. Kolter, ``Task-based end-to-end model learning in stochastic optimization,'' \emph{Advances in neural information processing systems}, vol.~30, 2017.

\bibitem{kotary_end_to_end_2021}
\BIBentryALTinterwordspacing
J.~Kotary, F.~Fioretto, P.~Van~Hentenryck, and B.~Wilder, ``End-to-{End} {Constrained} {Optimization} {Learning}: {A} {Survey},'' Mar. 2021, arXiv:2103.16378 [cs]. [Online]. Available: \url{http://arxiv.org/abs/2103.16378}
\BIBentrySTDinterwordspacing

\bibitem{elmachtoub_smart_2022}
\BIBentryALTinterwordspacing
A.~N. Elmachtoub and P.~Grigas, ``Smart “{Predict}, then {Optimize}”,'' \emph{Management Science}, vol.~68, no.~1, pp. 9--26, Jan. 2022, publisher: INFORMS. [Online]. Available: \url{https://pubsonline.informs.org/doi/10.1287/mnsc.2020.3922}
\BIBentrySTDinterwordspacing

\bibitem{fca}
\BIBentryALTinterwordspacing
{Financial Conduct Authority}, ``{FCA} handbook, specialist sourcebooks, conc 2 conduct of business standards: general,'' 2024, accessed on July 5, 2024. [Online]. Available: \url{https://www.handbook.fca.org.uk/handbook/CONC/2}
\BIBentrySTDinterwordspacing

\bibitem{civil_rights_div}
\BIBentryALTinterwordspacing
{Civil Rights Division, U.S. Department of Justice}, ``The equal credit opportunity act,'' 2024, accessed on July 5, 2024. [Online]. Available: \url{https://www.justice.gov/crt/equal-credit-opportunity-act-3}
\BIBentrySTDinterwordspacing

\end{thebibliography}

\end{document}